\documentclass[journal]{IEEEtran}
\usepackage{amsmath,amsfonts}
\usepackage{algorithmic}
\usepackage{algorithm}
\usepackage{array}
\usepackage[caption=false,font=normalsize,labelfont=sf,textfont=sf]{subfig}
\usepackage{textcomp}
\usepackage{stfloats}
\usepackage{url}
\usepackage{verbatim}
\usepackage{graphicx}
\usepackage{cite}
\usepackage{makecell}
\usepackage{multirow}
\usepackage{footnote}
\usepackage{calc} 
\begin{document}

\title{A Survey on Small Sample Imbalance Problem: Metrics, Feature Analysis, and Solutions}

\author{Shuxian Zhao, Jie Gui,~\IEEEmembership{Senior Member,~IEEE},  Minjing Dong, Baosheng Yu, Zhipeng Gui, ~\IEEEmembership{Member, ~IEEE}, Lu Dong, ~\IEEEmembership{Member, ~IEEE}, Yuan Yan Tang, ~\IEEEmembership{Life Fellow, ~IEEE}, and James Tin-Yau Kwok, ~\IEEEmembership{Fellow, ~IEEE}
\thanks{\textit{Corresponding author: Jie Gui.}}
\thanks{Shuxian Zhao and Lu Dong are with the School of Cyber Science and Engineering, Southeast University, Nanjing 210000,
China (e-mail: \protect\url{zhaosxian@seu.edu.cn}; ldong90@seu.edu.cn).}
\thanks{Jie Gui is with the School of Cyber Science and Engineering, Southeast University, Nanjing 211189, China, also with the Engineering Research Center of Blockchain Application, Supervision and Management, Southeast University, Nanjing 211189, China, and also with Purple Mountain Laboratories,
Nanjing 210000, China (e-mail: guijie@seu.edu.cn).}
\thanks{Minjing Dong is with the Department of Computer Science, City University of Hong Kong, Hong Kong (e-mail: minjdong@cityu.edu.hk).}
\thanks{Baosheng Yu is with the Lee Kong Chian School of Medicine at Nanyang Technological University, Singapore (email: baosheng.yu@ntu.edu.sg).}
\thanks{Zhipeng Gui is with the School of Remote Sensing and Information Engineering, Wuhan University, Wuhan 430079, China, and also with the Collaborative Innovation Center of Geospatial Technology, Wuhan University, Wuhan 430079, China (e-mail: zhipeng.gui@whu.edu.cn).}
\thanks{Yuan Yan Tang is with the Department of Computer and Information Science, University of Macau, Macau, China (e-mail: yytang@um.edu.mo).}
\thanks{James Tin-Yau Kwok is with the Department of Computer Science and Engineering, The Hong Kong University of Science and Technology, Hong Kong, China (e-mail: jamesk@cse.ust.hk).}
\thanks{A project associated with this survey has been created at \protect\url{https://github.com/guijiejie/Small-Sample-Imbalanced-Problem.}}}

\markboth{Journal of \LaTeX\ Class Files,~Vol.~14, No.~8, April~2025}%
{Shell \MakeLowercase{\textit{et al.}}: A Sample Article Using IEEEtran.cls for IEEE Journals}

\IEEEpubid{0000--0000/00\$00.00~\copyright~2025 IEEE}

\maketitle

\begin{abstract}
The small sample imbalance (S\&I) problem is a major challenge in machine learning and data analysis. It is characterized by a small number of samples and an imbalanced class distribution, which leads to poor model performance. In addition, indistinct inter-class feature distributions further complicate classification tasks. Existing methods often rely on algorithmic heuristics without sufficiently analyzing the underlying data characteristics. We argue that a detailed analysis from the data perspective is essential before developing an appropriate solution. Therefore, this paper proposes a systematic analytical framework for the S\&I problem. We first summarize imbalance metrics and complexity analysis methods, highlighting the need for interpretable benchmarks to characterize S\&I problems. Second, we review recent solutions for conventional, complexity-based, and extreme S\&I problems, revealing methodological differences in handling various data distributions. Our summary finds that resampling remains a widely adopted solution. However, we conduct experiments on binary and multiclass datasets, revealing that classifier performance differences significantly exceed the improvements achieved through resampling. Finally, this paper highlights open questions and discusses future trends.
\end{abstract}


\section{Introduction}
\IEEEPARstart{T}{he} continuous advancement of deep learning technologies has led to remarkable progress in tasks such as image classification, object detection, and image generation. However, sample collection remains challenging in industries such as steel, aerospace, and manufacturing product inspection, with fewer defect samples than normal. As a result, the issue of small sample imbalance (S\&I) \cite{ref30} is prevalent.

The small sample problem refers to limited training data, often leading to overfitting and poor generalization \cite{ref1,ref2}. To address this, researchers have proposed various approaches. For example, meta-learning \cite{ref3, ref4} trains models across multiple tasks to enable rapid adaptation to new tasks. Metric learning \cite{ref6,ref7} reduces reliance on large datasets by learning suitable distance metrics. Generative models such as Generative Adversarial Networks (GAN) \cite{ref8,ref9} and Variational Autoencoder (VAE) \cite{ref10} are used to generate new training samples. Transfer learning \cite{ref11,ref12} enhances learning efficiency on new tasks by transferring knowledge from pre-trained models. The class imbalance problem \cite{ref13, ref14, ref15} refers to unequal sample sizes between classes. A significant imbalance between majority and minority classes could encourage models to predict the majority class \cite{ref13, ref16}. Resampling techniques \cite{ref17} adjust the distribution of imbalanced data by oversampling \cite{ref18, ref19,ref20} the minority class or undersampling \cite{ref21, ref22} the majority class. Cost-sensitive learning \cite{ref23, ref24} enhances minority class learning by assigning different weights to different classes. Ensemble learning \cite{ref25, ref26} improves minority class accuracy by combining the predictions of multiple classifiers. Weighted loss functions \cite{ref27} and other deep learning methods \cite{ref28,ref29} are also used to address imbalance.

Existing solutions to the S\&I problem often focus on resampling or algorithmic adjustments. Most existing methods represent incremental improvements and lack comprehensive data analysis. However, the complexity of the dataset is a primary determinant of classification performance, which is exacerbated by class imbalance. This complexity may arise from data collection issues (e.g., noise, missing values) or characteristics intrinsic to the dataset (e.g., class overlap, lack of representative data, small disjuncts), all of which contribute to sample imbalance \cite{ref13, ref31, ref32}. Therefore, even with existing solutions in place, there remains a need for a new framework—one that can analyze these dataset complexities in detail and guide more effective solutions for S\&I problems.

Figure 1 provides a systematic analytical framework for addressing the S\&I problem.
\IEEEpubidadjcol
The framework begins by examining the key factors influencing classification performance and then identifies solutions. This approach transcends superficial concerns with sample size and class imbalance. We emphasize the fundamental value of quantifying the degree of imbalance and characterizing the internal features of the dataset. This deeper perspective exposes the limitations of relying solely on data augmentation or model tuning and provides a stronger foundation for S\&I research. Based on these ideas, this survey reviews recent solutions for each component of the framework, outlines unresolved challenges, and proposes future directions.

\begin{figure*}[!t]
	\centering
	\includegraphics[width=\linewidth]{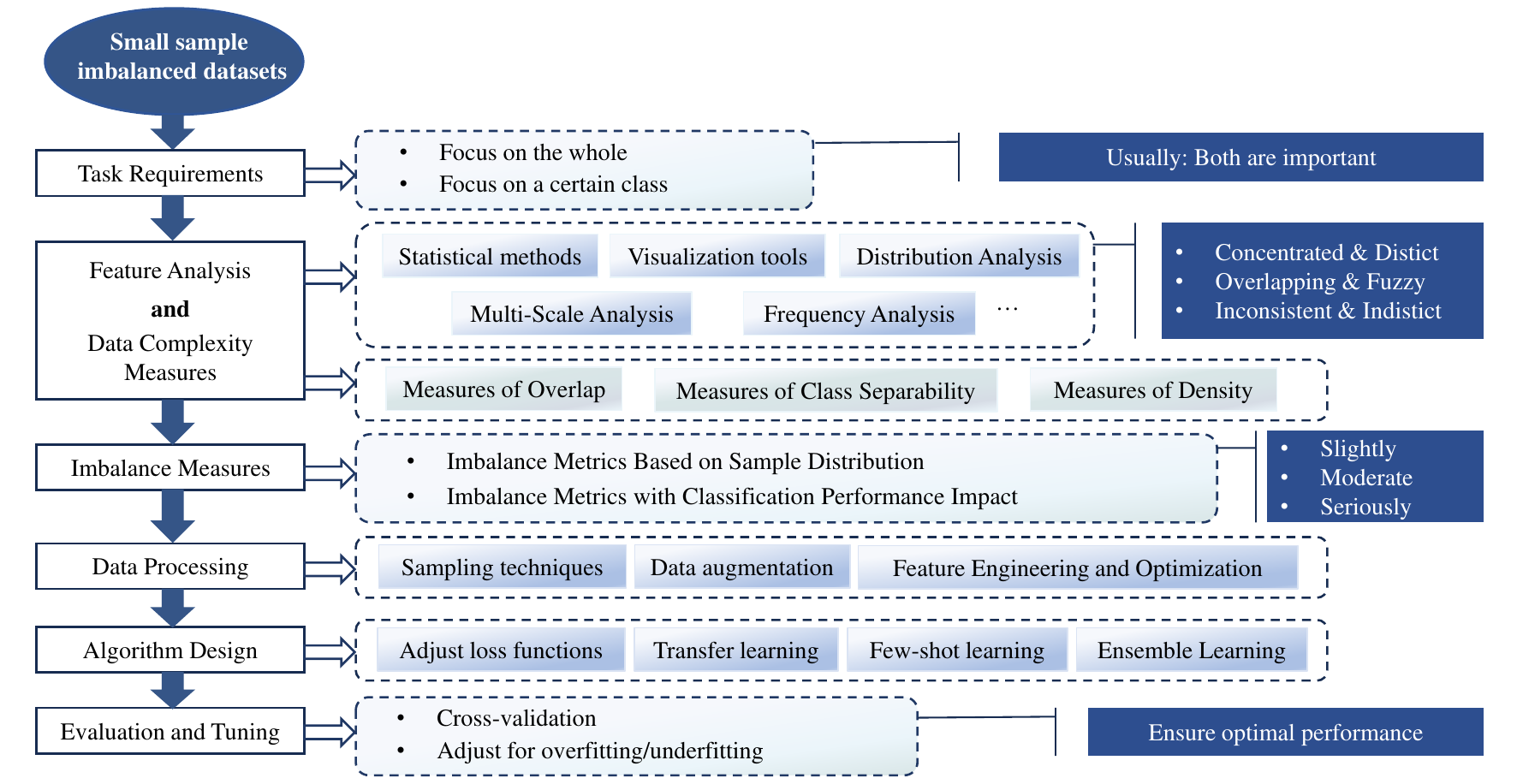}
	\caption{The Systematic Analysis Framework of the Small Sample Imbalance Problem.}
	\label{fig_1}
\end{figure*}

The contributions of this paper and the differences from previous work can be summarized as:

1)	This paper defines the S\&I problem, explores its current state, and analyzes its multidimensional factors. We summarize imbalance measurement and data complexity assessment methods. These methods quantify inter-class differences and intra-class distributions, providing practical guidance for solution design.

2)	Unlike traditional reviews focused on data, features, and algorithms, we examine S\&I problems through sample and feature distributions to better uncover the roots of learning difficulty. We summarize recent solutions across conventional, complexity-based, and extreme scenarios, offering a more diagnostic perspective on the S\&I problem.

3)	Our experiments on resampling methods demonstrate that classifier performance dominates, with resampling adaptation playing a secondary role. Furthermore, algorithm capability and dataset characteristics need to be carefully balanced, which provides valuable insights into future algorithm designs for S\&I problems.

The rest of this paper is organized as follows. Section II defines S\&I, outlines its existence situation. Section III summarizes the dataset metrics and analysis methods, and also introduces the datasets and evaluation methods. Section IV outlines three solutions and presents comparative experiments on resampling methods. Section V discusses open problems and future research directions. Finally, Section VI concludes the paper.

\section{Small Sample Imbalance Problem}
The S\&I problem is particularly challenging because the data are limited, and every sample is crucial. Any bias can significantly impact the model's performance. Therefore, we first define and analyze this problem.

\subsection{Small Sample and Few Shot Learning} 
``Small sample'' usually refers to a situation where the number of samples taken from the whole population is relatively small \cite{ref33}. It is widely used in statistics and data analysis to describe dataset size without addressing how to leverage it for learning. The small sample problem often arises when data acquisition is difficult or costly. In machine learning, it can lead to overfitting. Models struggle to learn robust patterns from limited data, leading to discrepancies between training and testing performance. Few-shot learning \cite{ref34} is a term in machine learning that refers to a model's ability to generalize tasks from very small samples. Commonly used methods include transfer learning and meta-learning. These methods aim to enable models to quickly adapt to new tasks or classes without relying on large amounts of data. While both terms relate to sample size, ``small sample'' describes the dataset itself, whereas ``few-shot learning'' refers to strategies for learning effectively in such scenarios.

\subsection{S\&I Dataset Define} 
We are given a dataset $D$ containing $N$ samples, $D=\{(x_i,y_i)\}_{i=1}^N$, where ${{x}_{i}}$ represents the feature of the $i$th sample and ${{y}_{i}}$ is the corresponding class label. $C$ be the total number of classes. The class ${{c}_{j}}$ (where $j=1,2,\cdots, C$) contains ${{N}_{j}}$ samples. $N_j$ is defined as $N_j=\sum_{i=1}^N\mathbb{I}(y_i=c_j)$. $\mathbb{I}\left( \cdot \right)$ is the indicator function, which takes the value 1 when the condition inside parentheses is satisfied and 0 otherwise.

The small sample condition occurs when the total number of samples, $N$, is insufficient for effective generalization. This is expressed as $N\ll M$, where $M$ is the standard dataset size for the application. The imbalance condition requires that at least one class ${{c}_{j}}$ has a sample ratio $\frac{{{N}_{j}}}{N}$ significantly smaller than $\frac{{{N}_{k}}}{N}$ for all $k$ not equal to $j$.

Therefore, a dataset $D$ is considered an S\&I problem if it satisfies the formula \eqref{deqn_ex1}.

\begin{equation}
	\label{deqn_ex1}
	D=\left\{ \{({{x}_{i}},{{y}_{i}})\}_{i=1}^{N}\left| N\ll M\land \exists{j}, \frac{{{N}_{j}}}{N}\ll \frac{{{N}_{k}}}{N} \right. \right\}.
\end{equation}

\subsection{Overview of S\&I} 
\begin{figure}[!t]
	\centering
	\includegraphics[width=\linewidth]{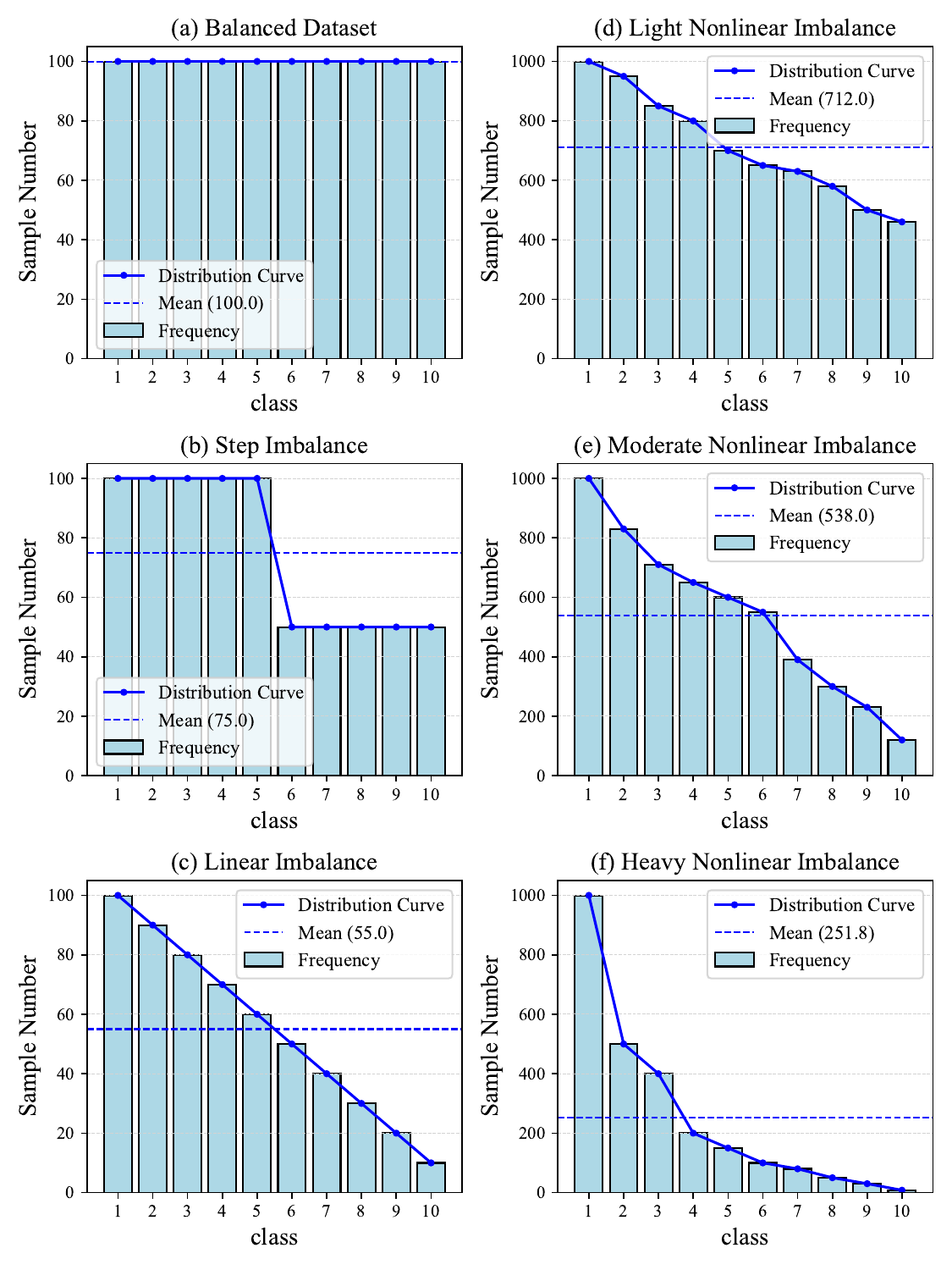}
	\caption{Distribution of inter-class imbalance.}
	\label{fig_2}
\end{figure}

\textit{1) Inter-class sample imbalance:} Class imbalance in small sample datasets typically refers to the distribution of inter-class imbalances. Figure 2 shows that data imbalance can be classified into three main types: step imbalance, linear imbalance, and nonlinear imbalance \cite{ref14, ref31}.  In addition, the degree of imbalance can be categorized from light to heavy. Figures (c), (e), and (f) further distinguish several scenarios: an even split between classes with many and few samples, a majority of classes with many samples, and a majority of classes with few samples. In real-world applications, the diversity in the distribution of sample sizes among classes complicates the issue of class imbalance. Therefore, it is relevant to investigate solutions for different sample size differences and their extent.

\textit{2) Other imbalances:} The class imbalance issue may not be problematic, as it does not inherently correlate with classification complexity \cite{ref35}. We focus on examining whether the factors influencing classification performance are truly due to this imbalance. In addition to class imbalance, small sample datasets may exhibit other imbalances.

As shown in Figure 3, feature imbalance refers to the imbalanced distribution of specific feature values. Some features may appear infrequently in certain categories, making it difficult for the model to learn the relevant features effectively. Alternatively, certain features may dominate in some samples while being less apparent in others, which affects the model's ability to capture the full range of features. This type of imbalance is associated with small disjuncts, rare sub-concepts and instances, and outliers \cite{ref36, ref37, ref38}. Sample quality imbalance may arise from differences between high-quality and low-quality samples. This issue is particularly pronounced when samples contain noise \cite{ref39}, which impedes the model learning process. Temporal imbalance \cite{ref40} in time-series data manifests as differences in sample quantity and category distribution across various time periods, potentially leading to reduced model performance in specific time frames. Spatial imbalance \cite{ref41} occurs in geographical or spatial data, characterized by uneven sample quantities and class distributions across regions. This regional disparity can compromise a model's generalization capability in specific areas. Each type of imbalance can have varying effects on model performance and generalization, contributing to the complexity of S\&I issues. Therefore, when addressing S\&I problems, it is crucial to choose appropriate strategies or methods based on the specific situation.

\begin{figure*}[!t]
	\centering
	\includegraphics[width=\linewidth]{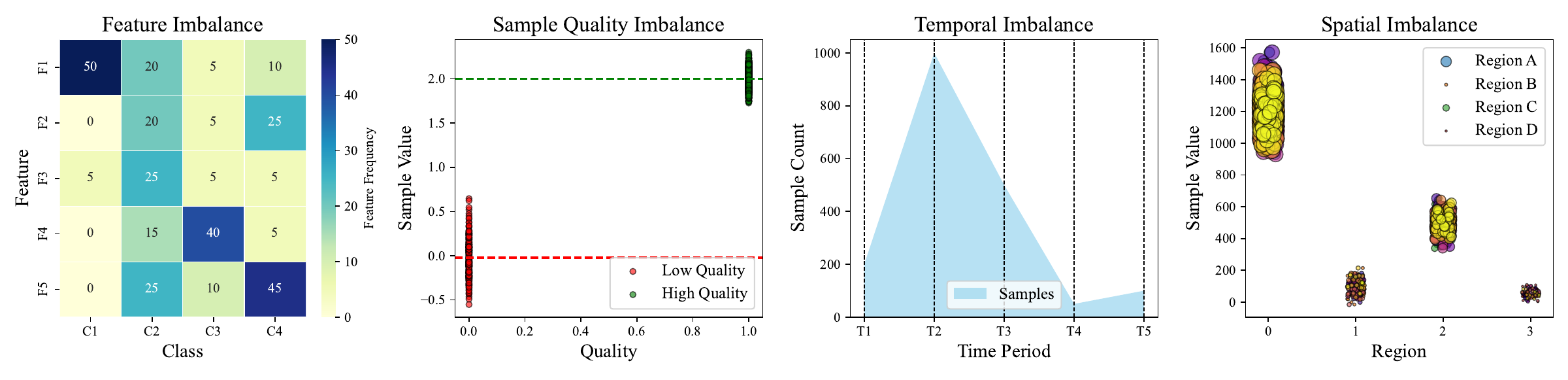}
	\caption{Other Example Imbalances.}
	\label{fig_3}
\end{figure*}

Although various types of imbalances may take different forms, they often manifest as imbalanced feature distributions during the feature learning stage. Therefore, these imbalance problems from different sources can be consistently interpreted from the perspective of feature imbalance. Building on this perspective, the following discussion focuses on class distribution imbalance, feature properties, and data complexity.

\section{S\&I Measurement and Analysis} 
In S\&I datasets, the degree of class imbalance can affect model performance differently. When feature distributions are clearly separated, models may still perform well despite imbalance. In contrast, significant overlap between class features can worsen misclassification \cite{ref36, ref37}. Therefore, it is essential to identify the nature of the problem first to address it more accurately. Imbalance measurement and feature analysis can reveal class differences and intra-class distributions, thereby providing a guide for algorithm design and model evaluation.

\subsection{Imbalance measurement}
In different scenarios, appropriate imbalance metrics can help increase the sample size through data augmentation and inform algorithm design. This section reviews multiple imbalance assessment metrics. These include metrics that focus solely on the distribution of sample numbers, as well as those that consider the impact of imbalance on classification performance.

\textit{1) Imbalance Metrics Based on Sample Distribution:} These metrics assess the imbalance in a dataset by measuring the ratio of sample sizes between classes.

The imbalance ratio (IR) is defined as:
\begin{equation}
	\label{deqn_ex2}
	IR=\frac{{{N}_{\max }}}{{{N}_{\min }}},
\end{equation}
where ${{N}_{\min }}$ and ${{N}_{\max }}$ represent the sample numbers of the minimum and maximum classes, respectively. A larger IR indicates a higher degree of imbalance. If the data are well-separated, IR can reflect the degree of multi-class imbalance but may overlook detailed class distribution information.

The Gini Index \cite{ref42} is a statistical measure commonly used to represent income or wealth inequality. In classification tasks, it assesses the purity or impurity of a dataset, defined as
\begin{equation}
	\label{deqn_ex3}
	G=1-\sum_{i=1}^C\left(\frac{N_i}N\right)^2.
\end{equation}
The Gini index ranges from 0 to 1. A smaller value indicates a higher purity of the dataset, meaning most samples belong to a single class. Conversely, a larger value reflects greater impurity, suggesting a more balanced distribution of classes. It is commonly used in the CART decision tree algorithm \cite{ref43} to select the optimal splitting feature, aiming to minimize the Gini index in the resulting subsets.

The Imbalance-degree (ID) \cite{ref44} is a normalized metric for a multi-class dataset. It measures the class imbalance by considering the distance between the sample distribution $\zeta =\left( \frac{{{N}_{1}}}{N},\frac{{{N}_{2}}}{N},\cdots,\frac{{{N}_{C}}}{N} \right)$ and the ideal balanced distribution $\text{e}=\left( \frac{1}{C},\frac{1}{C},\cdots,\frac{1}{C} \right)$. It is defined as:
\begin{equation}
	\label{deqn_ex4}
	ID=\frac{{{d}_{\Delta }}\left( \zeta ,\text{e} \right)}{{{d}_{\Delta }}\left( {{\iota }_{m}},\text{e} \right)}+\left( m-1 \right).
\end{equation}
Here, $m$ represents the number of minority classes. ${{d}_{\Delta }}$ is a distance similarity function (e.g., Euclidean distance, Chebyshev distance). ${{\iota }_{m}}$ is a distribution with $m$ minority classes that is maximally distant from the balanced distribution $\text{e}$. In a perfectly balanced scenario, the ID value is zero. Otherwise, it lies within the $(m-1,m]$ range. The choice of distance similarity function influences the ID value, and improper use may harm the results. Additionally, this method finds that the more minority classes in the dataset, the higher the degree of class imbalance. This is incorrect \cite{ref45}. For example, the class frequencies of two datasets are (1,1000,1000) and (1000,1000,1003). The first dataset is imbalanced, while the second one is almost balanced. However, the second one has a higher ID due to having two minority classes.

Likelihood Ratio Imbalance (LRID) \cite{ref45} measures the difference between imbalanced and perfectly balanced class distributions through a likelihood-ratio test, defined as \eqref{deqn_ex5}. This method addresses the issues present in ID.
\begin{equation}
	\label{deqn_ex5}
	\text{LRID}=-2\sum\limits_{i=1}^{C}{{{N}_{i}}\ln \frac{N}{C{{N}_{i}}}}.
\end{equation}

The Imbalance Coefficient (IC) \cite{ref46} is designed for binary classification problems, defined as \eqref{deqn_ex6}. Its value, $\delta$, ranges from [-1, 1], with a value of 0 representing perfect balance. This symmetry and bounded range make $\delta$ more convenient and intuitive for mathematical handling and interpretation.
\begin{equation}
	\label{deqn_ex6}
	\delta =2\frac{{{N}_{0}}}{N}-1,
\end{equation}
where ${{N}_{0}}$ represents the number of samples in the first class.

Relative Imbalance Rate (RIR) \cite{ref47}: This method amplifies minority class importance by weighting its frequency with the RIR during classification.
\begin{equation}
	\label{deqn_ex7}
    RIR\left( {{c}_{i}} \right)=\frac{{{N}_{\max }}}{{{N}_{i}}},
\end{equation}
where ${N}_{\max}$ is the number of samples in the majority class and ${N}_{i}$ is the number of samples in class $c_i$.

\textit{2) Measuring Imbalance with Classification Performance:} Some studies suggest that an ideal metric should negatively correlate with classification performance \cite{ref45}. More refined imbalance metrics better capture imbalance effects, thereby promoting the development of more effective solutions.

The Augmented R-value \cite{ref48}, an improvement of the R-value \cite{ref49}, estimates the degree of overlap in binary classification data while accounting for class imbalance. Since overlap significantly impacts classifier performance more than imbalance, the calculation of $R_{avg}$ shows a strong negative correlation with classifier performance.
\begin{equation}
	\label{deqn_ex8}
	R\left( {{c}_{i}} \right)=\frac{1}{{{N}_{i}}}\sum\limits_{m=1}^{{{N}_{i}}}{\mathbb{I}\left( \left| kNN\left( {{P}_{{{c}_{i}},m}},D-{{U}_{{{c}_{i}}}} \right) \right|-\theta  \right)},
\end{equation}
\begin{equation}
	\label{deqn_ex9}
	{{R}_{aug}}=\frac{1}{IR+1}\left( R\left( {{c}_{\min }} \right)+IR\cdot R\left( {{c}_{\text{maj}}} \right) \right),
\end{equation}
where ${{U}_{{{c}_{i}}}}$ denotes the set of instances of class ${{c}_{i}}$, and $D$ is the set of all instances. ${{P}_{{{c}_{i}},m}}$ represents the $m$-th instance of class ${{c}_{i}}$. $kNN(P,S)$ refers to the $k$ nearest neighbors of instance $P$ within the set $S$. $\theta$ is a threshold. $\mathbb{I}\left( x \right)$ is an indicator function, which is 1 if $x>0$ and 0 otherwise.

Zhu et al. \cite{ref50} demonstrated that using the same IR to describe imbalance in datasets with different dimensions is inappropriate. Therefore, they introduced an adjusted IR for binary classification. The metric contains a penalty term based on the number of discriminatory features, which are determined by a Pearson correlation test.
\begin{equation}
	\label{deqn_ex10}
	\text{adjustedIR=}IR-\lambda \log \left( {{p}^{*}} \right),
\end{equation}
where ${p}^{*}$ represents the number of discriminative features determined by the Pearson correlation test, and $\lambda$ is the parameter controlling the importance of the penalty term. The penalty term, $\log \left( {{p}^{*}} \right)$, adjusts the impact of dimensionality on classification performance. The adjusted IR decreases as the number of discriminative features increases.

The Individual Bayes Imbalance Impact Index ($\text{IBI}^{3}$) \cite{ref51} measures prediction score differences for minority samples across balanced and imbalanced conditions. This quantifies the extent to which individual samples are affected by class imbalance. The Bayes Imbalance Impact Index ($\text{BI}^{3}$) is the average of $\text{IBI}^{3}$ values for all minority class samples, reflecting the overall impact of imbalance on the dataset. It applies only to binary classification. $\text{BI}^{3}$ can be used to assess whether imbalance mitigation techniques, such as sampling or cost-sensitive methods, are needed to improve classifier performance. The higher the $\text{BI}^{3}$ value, the greater the potential to improve performance through these techniques.
\begin{equation}
	\label{deqn_ex11}
	\text{IBI}^{3}\left( {{x}_{i}} \right)=\left\{ \begin{matrix}
		\frac{IR}{M+IR}-\frac{1}{M+1} & M={{k}_{0}}  \\
		\frac{IR\left( {{k}_{0}}-M \right)}{M+IR\left( {{k}_{0}}-M \right)}-\frac{{{k}_{0}}-M}{{{k}_{0}}} & \text{Otherwise} \\
	\end{matrix} \right.,
\end{equation}
\begin{equation}
	\label{deqn_ex12}
	\text{BI}^{3}=\frac{1}{{{N}_{\min }}}\sum\limits_{\begin{smallmatrix} 
		\left( {{x}_{i}},{{y}_{i}} \right)\in D \\ 
		{{y}_{i}}={{c}_{\min }} 
\end{smallmatrix}}{\text{IB}{{\text{I}}^{3}}\left( {{x}_{i}} \right)},
\end{equation}
where ${{x}_{i}}$ represents an instance from the minority class, and $M$ denotes the number of majority class instances among the ${{k}_{0}}$-nearest neighbors of ${{x}_{i}}$.

The Imbalance Factor (IF) \cite{ref52} is a new scale proposed to measure based on information theory and Rényi entropy. It works for binary and multi-class datasets, outputs values within a fixed range of [0, 1], and strongly correlates with classifier performance.
\begin{equation}
	\label{deqn_ex13}
	IF=\frac{\frac{1}{1-\alpha }{{\log }_{2}}\left( \sum\nolimits_{i=1}^{C}{{{\left( \frac{{{N}_{i}}}{N} \right)}^{\alpha }}} \right)}{{{\log }_{2}}\left( C \right)},
\end{equation}
where $\alpha$ is the order of Rényi entropy, which can be chosen to adjust the sensitivity of the scale. By setting $\alpha$ to 1, 2, and $\infty$, the special versions of $IF_{Shannon}$, $IF_{Collision}$, and $IF_{Min}$ are obtained, respectively.

Lin et al. \cite{ref53} proposed a purposive data augmentation strategy based on minority class imbalance Ratio (MiCIR). This strategy introduces the multiclass imbalance ratio (MIR) and the MiCIR.
\begin{equation}
	\label{deqn_ex14}
	\operatorname{MIR}=\frac{{{N}_{m}}}{{{N}_{l}}} \text{,} \quad MiCIR\left( {{c}_{i}} \right)=\frac{{{N}_{\max }}}{{{N}_{i}}}\left( {{N}_{i}}<\frac{N}{C} \right).
\end{equation}

The majority and minority classes in a multiclass dataset are divided using the ratio $\frac{N}{C}$. ${{N}_{m}}$ and ${{N}_{l}}$ represent the total number of samples in the majority and minority classes, respectively. The purpose of MIR and MiCIR in the paper is to identify the truly insufficient classes in the dataset, which are then placed into the to-be-augmented set ${{D}_{TBA}}$.
\begin{equation}
	\label{deqn_ex15}
	{{D}_{TBA}}=\left\{ \begin{matrix}
	{{D}_{l}} & \text{MIR}\le 1  \\
	\left\{ {{c}_{i}}\in {{D}_{l}}|\text{MiCIR}({{c}_{i}})>\text{MIR} \right\} & \text{MIR}>1  \\
\end{matrix} \right.,
\end{equation}
where ${{D}_{l}}$ denotes the set of all minority classes. This study also conducts a preliminary experiment to identify easily misclassified classes. Considering both the small sample size and high misclassification rate determines which classes require data augmentation. This method effectively assesses the need for data augmentation and identifies target classes.

\textit{Summary:} Figure 4 presents the correlation heatmaps of imbalance metrics across all binary and multi-class datasets listed in Table III. Color intensity indicates the strength of correlation. IR, ID, and LRID show strong positive correlations in both tasks, suggesting consistency and generality in measuring data imbalance. IF is negatively correlated with most metrics, indicating an opposite measurement direction. Gini tends to be positively correlated with other metrics in multi-class tasks but negatively correlated in binary tasks, indicating its sensitivity varies across task types. $R_{avg}$ has minimal correlation with other metrics, highlighting the uniqueness of its inclusion of class overlap calculations.

\begin{figure}[!t]
\centering
\includegraphics[width=\linewidth]{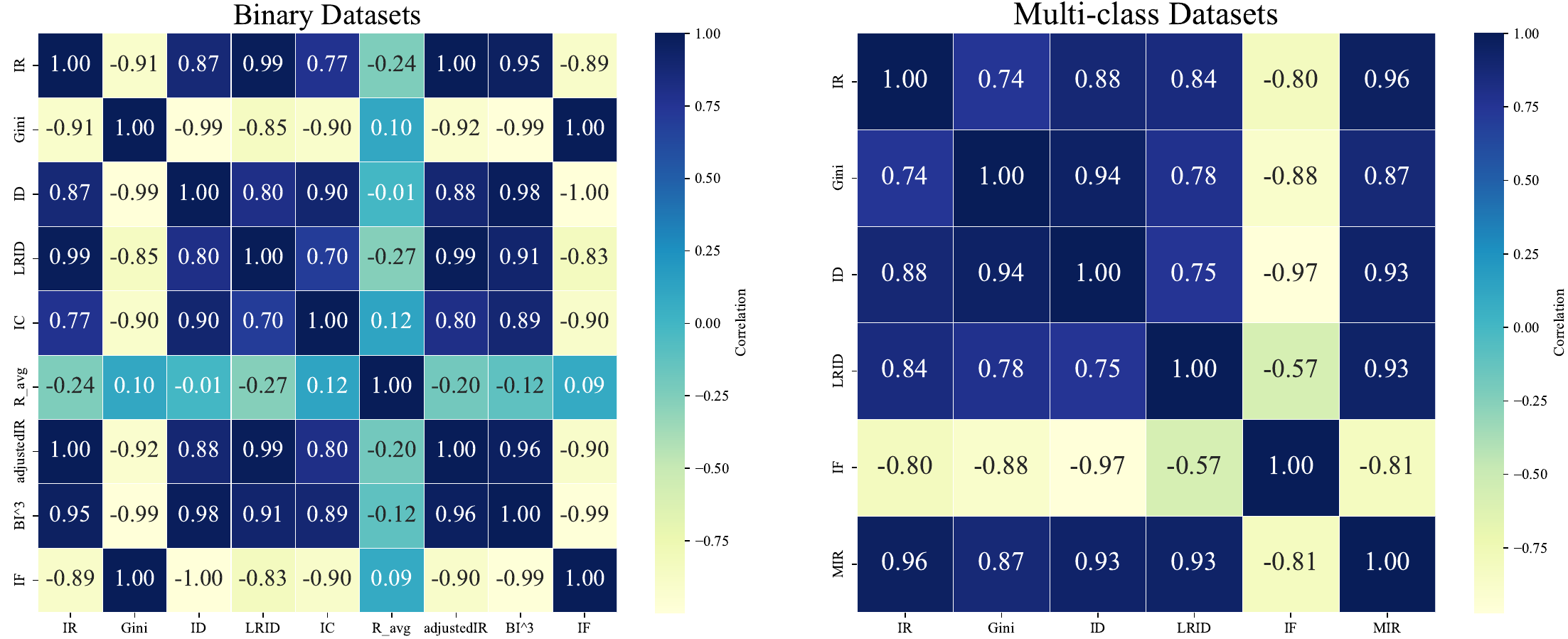}
\caption{Correlation between imbalance indicators. (In adjustedIR, $\lambda$ is set to 1. In IF, $\alpha$ is set to 2. In $R_{\text{aug}}$, $k$ is 5 and $\theta$ is 2. In $\text{IBI}^{3}$, $k$ is 5.)}
\label{fig_10}
\end{figure}

\subsection{Feature Analysis and Data Complexity Measurement} 
In S\&I image classification, imbalanced sample distribution is just one factor influencing performance and may not be the primary cause of degradation. Some classes may be overshadowed or difficult to distinguish in feature space \cite{ref38, ref54}. It is also crucial to analyze the dataset characteristics. One can identify easily recognizable classes and those requiring special handling by leveraging statistical analysis, dimensionality reduction, and distribution visualization. This analysis facilitates the selection of suitable preprocessing and feature engineering techniques. It also accelerates model development for real-world imbalanced data scenarios \cite{ref38}. Table I summarizes five common types of feature distribution analysis.

\begin{table*}
	\begin{center}
		\caption{Feature Analysis Methods.}
		\label{tab1}
		\begin{tabular}{ l l l }
			\hline
			\textbf{Type} & \textbf{Method} & 	\textbf{Application Objective} \\
			\hline
			Statistical Analysis &	Mean and Variance& \makecell[l]{ Assess feature central tendency and dispersion for image contrast and brightness.}\\ 
			& Skewness and Kurtosis	& \makecell[l]{Evaluate distribution symmetry and heavy-tailedness to identify class imbalance and outliers.}\\
			& Shannon Entropy & \makecell[l]{Measure feature complexity and information content, aiding classification difficulty evaluation.}\\
			\hline 
			\multirow{3}{*}{\makecell[l]{Dimensionality \\Reduction and\\ Visualization}}   & t-SNE & \makecell[l]{Visualize high-dimensional data to assess class separability in lower-dimensional space.}\\
			& PCA & \makecell[l]{Reduce dimensionality and highlight key features to enhance classification performance.}\\
			& LDA & \makecell[l]{Maximize class separability by reducing within-class variance, improving classifier discrimination.}\\
			\hline
			Distribution Analysis &	KL Divergence & 	\makecell[l]{Measure class distribution differences to identify discriminative features.}\\
			& KS Test &	\makecell[l]{Test distribution differences between classes, particularly in imbalanced datasets.}\\
			\hline
			Multi-Scale Analysis & Wavelet Transform &  \makecell[l]{Extract multi-resolution features for texture analysis and multi-scale data problems.}\\
			\hline
			Frequency Analysis & Fourier Transform & 	\makecell[l]{Extract frequency-domain features for periodic pattern recognition in images.}\\
			\hline 
		\end{tabular}
	\end{center}
\end{table*}

The t-SNE plot visualizes intra-class and inter-class distributions by projecting high-dimensional data into a lower-dimensional space. In preliminary analysis, it helps assess data separability. Fig.5 (a) large inter-class distances and compact intra-class distributions indicate high separability, making classification relatively easier. After model training, t-SNE can also evaluate model adaptability by revealing the clarity of classification results. Well-defined class boundaries suggest that the model effectively captures data features. Figure 5 presents t-SNE visualizations of three distinct feature distributions. Based on these distributions, we propose different strategies for handling S\&I in the following sections.

\begin{figure*}[!t]
	\centering
	\includegraphics[width=\linewidth]{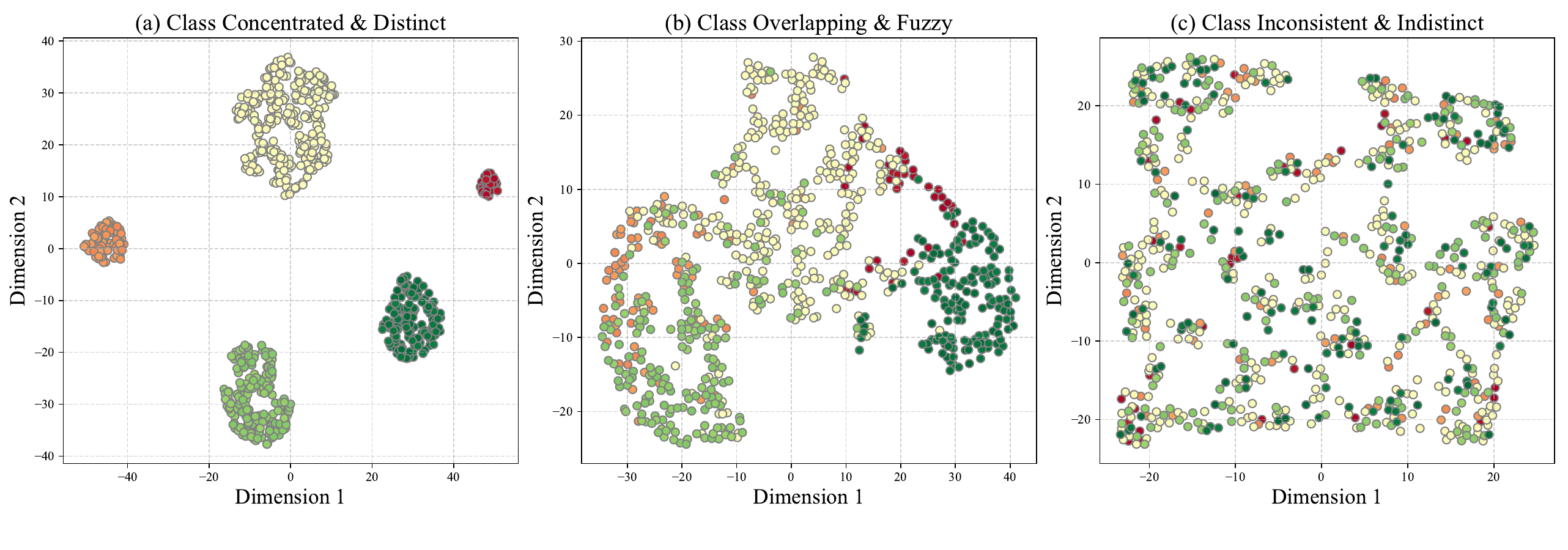}
	\caption{t-SNE plots of three different feature distributions.}
	\label{fig_4}
\end{figure*}

Feature-based data complexity measures serve as assessment tools for dataset classification difficulty. They offer insights into how data characteristics affect machine learning model performance. Table II shows these measures analyze data from multiple perspectives, including feature overlapping measures, neighborhood measures, linear separability measures, and other complexity measures for imbalanced datasets. Quantifying dataset complexity assists researchers in model selection and optimization strategy development. These capabilities prove particularly valuable when addressing class imbalance, overlapping distributions, or high-dimensional data challenges. Detailed descriptions, formulations, and references for these measures can be found in \cite{ref55, ref56, ref57, ref58, ref59, ref60}.

\begin{table}
\begin{center}
\caption{Data Complexity Measures.}
\label{tab2}
\begin{tabular}{ l  l }
\hline
\textbf{Type} & \textbf{Measures} \\
\hline
Feature Overlapping Measures & F1, F1v, F2, F3, F4, IN \\
Neighborhood Measures &	N1, N2, N3, N4, T1, LSC \\
Linear Separability Measures &	L1, L2, L3\\
\makecell[l]{Other complexity measures  for\\ imbalanced dataset}  &	CM, wCM, dwCM\\
\hline 
\end{tabular}
\end{center}
\end{table}

\subsection{Dataset} 
The UCI Machine Learning Repository (\url{https://archive.ics.uci.edu/datasets}) and KEEL Data Repository\cite{ref61} (\url{https://sci2s.ugr.es/keel/imbalanced.php}) are key resources for machine learning and data mining research. The UCI repository offers diverse datasets for classification, regression, and clustering. KEEL specializes in datasets for evaluating evolutionary algorithms and optimization techniques. Table III summarizes the datasets commonly used in the studies referenced \cite{ref62, ref63, ref64, ref65, ref66}. In some binary classification tasks, datasets originally designed for multi-class classification are often restructured into binary classification problems for experimental purposes. We calculate the above imbalance metrics corresponding to each dataset in Table III to better understand the distribution.

\begin{table*}
\begin{center}
\caption{Examples of Commonly Used Datasets.}
\label{tab3}
\begin{tabular}{l|ccccc|cccc}
\hline   
\textbf{Task}& \multicolumn{5}{c|}{\textbf{Binary}} & \multicolumn{4}{c}{\textbf{Multi-class}} \\
\hline   
Dataset& Breast-Cancer& ecol& Wine& yeast& Glass& ecoli& Wine& yeast& Glass  \\
\hline
Size& 569& 336& 178& 1484& 214& 336& 178& 1484& 214 \\
Features& 30& 7& 13& 8& 9& 7& 13& 8& 9 \\
Class& 2& 2& 2& 2& 2& 8& 3& 10& 6     \\
Class Distribution& 357/212& 35/301 & 71/107 & 51/1433 & 70/144 & \makecell[c]{143/77/52/\\20/5/2/2} & 71/59/48 & \makecell[c]{463/429/244/163/\\51/44/35/30/20/5}& \makecell[c]{76/70/29/\\17/13/9} \\
IR& 1.68& 8.6& 1.51& 28.1& 2.06& 71.5& 1.48& 92.6& 8.44 \\
Gini& 0.48& 0.19& 0.48& 0.07& 0.44& 0.73& 0.66& 0.78& 0.74 \\
ID& 0.25& 0.79   & 0.2 & 0.93    & 0.35   & 5.4   & 1.22  & 6.36& 4.08       \\
LRID  & 37.36   & 241.25 & 7.33   & 1613.23 & 26.12  & 377.88   & 4.48 & 1710.63     & 121.17  \\
IC& -0.25   & 0.79   & 0.2    & 0.93    & 0.35   & -   & -        & -   & -     \\
Augmented R-value& 0.02   & 0.06   & 0      & 0.02    & 0.14   & -    & -   & -  & -      \\
adjusted IR & -1.53   & 7.21   & -0.69  & 27  & 0.67   & -     & -    & -      & -        \\
$\text{IBI}^{3}$ & 0.02 & 0.3    & 0.02   & 0.51    & 0.07   & -    & -     & -     & -      \\
IF   & 0.91   & 0.3    & 0.94   & 0.1     & 0.84   & 0.63     & 0.98     & 0.65   & 0.74   \\
MIR  & -  & -   & -      & -  & -   & 4.25 & 0.66     & 7.02     & 2.15    \\
Complexity   & +    & +++    & +  & +++++   & ++++   & ++++      & ++  & +++++     & +++++     \\
\hline   
\end{tabular}
\end{center} 
 \hspace{0.5cm}
 \footnotesize \textit{Complexity is evaluated based on \cite{ref62}, \cite{ref66}, with more ``+'' symbols indicating higher dataset complexity. ``-'' indicates not applicable to the task.}
\end{table*}

\subsection{Evaluation Metrics} 
It is critical to select the appropriate evaluation metric according to different needs when dealing with S\&I datasets. Individual metrics can be biased due to data imbalance. Table IV summarizes several common evaluation metrics \cite{ref67, ref68, ref69, ref70}. Macro-average Recall, Macro-average Precision, and Macro-average F1-score are commonly used in multi-class imbalanced classification. The ROC curve shows the relationship between the true positive rate (TPR) and the false positive rate (FPR) at different thresholds. The area under the curve (AUC) is a comprehensive performance measure. The precision-recall curve (PRC) outperforms the ROC curve for imbalanced datasets. This is because the ROC curve tends to be overly optimistic as it is insensitive to class imbalance. In contrast, the PRC focuses on the minority class. However, AUC remains widely used. The debate over the suitability of the Matthews correlation coefficient (MCC) for imbalanced datasets remains inconclusive. While Zhu \cite{ref71} claims that MCC's nonlinearity and sensitivity to class size ratios make it unsuitable for imbalanced data, others \cite{ref72, ref73} argue that MCC can still be useful. This highlights the complexity of selecting evaluation metrics for imbalanced datasets. Different metrics may have varying degrees of effectiveness depending on the data complexity and the analysis goals.

Ren et al. \cite{ref33} found that test accuracy alone cannot precisely reflect the behavioral changes of models under imbalanced data, particularly regarding convergence speed and correctness. To address the limitations of traditional evaluation metrics, they introduced a new metric, convergence speed, to provide a more comprehensive description of model behavior. Assuming that the convergence condition for model training is when the training loss reaches a certain threshold, the convergence indicator $\delta$ is defined as follows:
\begin{equation}
	\label{deqn_ex16}
	\delta =\sqrt{\frac{\sum\nolimits_{i=1}^{C}{{{\left( 1-{{{\tilde{L}}}_{i}} \right)}^{2}}\cdot \varepsilon }}{C\cdot S}},
\end{equation}
where $\varepsilon$ denotes the tolerance coefficient, controlling the strictness of convergence; $C$ represents the number of classes; $S$ is the number of samples in the training subset; and ${{\tilde{L}}_{i}}$ is the predicted label value of the model. If the model's training loss exceeds $\delta$ within the specified number of iterations, it is deemed not to have converged. The more non-convergent trials there are, the slower the convergence speed of the model will be. This convergence indicator offers a novel perspective on investigating the impact of imbalanced data on models.

\begin{table}
	\begin{center}
		\caption{Evaluation Metrics for S\&I Classifications.}
		\label{tab4}
		\begin{tabular}{ lc }
			\hline
			\textbf{Evaluation metric}&	\textbf{Formula}\\
			\hline 
			Confusion Matrix& $\left( \begin{matrix}
				\text{True Positives}\left( TP \right) & \text{False Negatives}\left( FN \right)  \\
				\text{False Positives}\left( FP \right) & \text{True Negatives}\left( TN \right)  \\
			\end{matrix} \right)$ \\[8pt]
			Accuracy&	$\displaystyle \frac{TP+TN}{TP+FN+FP+TN}$ \\[8pt]
			Precision&	$\displaystyle \frac{TP}{TP+FP}$\\[8pt]
			Recall,TPR&	$\displaystyle \frac{TP}{TP+FN}$\\[8pt]
                FPR& $\displaystyle \frac{FP}{FP+TN}$\\[8pt]
			F1-score&	$\displaystyle \frac{2\cdot Precision\cdot Recall}{Precision+Recall}$ \\[8pt]
			Weighted F1-score&	$\displaystyle \sum\nolimits_{i}^{{}}{{{w}_{i}}\cdot F{{1}_{i}}}$\\[6pt]
			AUC-ROC&	\makecell[c]{$\int_{0}^{1}{TPR\left( FPR \right)dFPR}$} \\[6pt]
			MCC & $ \frac{TP\cdot TN-FP\cdot FN}{\sqrt{(TP+FP)(TP+FN)(TN+FP)(TN+FN)}}$\\[8pt]
			G-Mean & $\displaystyle \sqrt{{{Recall}_{+}}\cdot {{Recall}_{-}}}$ \\
			\hline
		\end{tabular}
	\end{center} 
	\footnotesize \textit{The formulas are for binary classification but can be extended to multi-class tasks.}
\end{table}

\section{Solutions and Experimental}
Effectively addressing the S\&I problem requires selecting appropriate models and data processing methods based on dataset characteristics to develop robust solutions. As shown in Fig. 5(a), when intra-class features are concentrated and inter-class differences are distinct, strong classification performance is often achieved without additional data processing. In contrast, Fig. 5 (b) depicts a more challenging scenario where inter-class features overlap and boundaries are fuzzy, which are common issues in real-world applications. Fig. 5 (c) illustrates an extreme case with substantial intra-class variation, making class separation nearly impossible in both binary and multi-class classification. Based on these insights, we categorize three types of solutions: conventional S\&I solutions, solutions for data complexity, and solutions for extreme S\&I. Figure 6 illustrates the emphasis on different levels of solutions within these three categories and mainstream method types.

\begin{figure*}[!t]
	\centering
	\includegraphics[width=\linewidth]{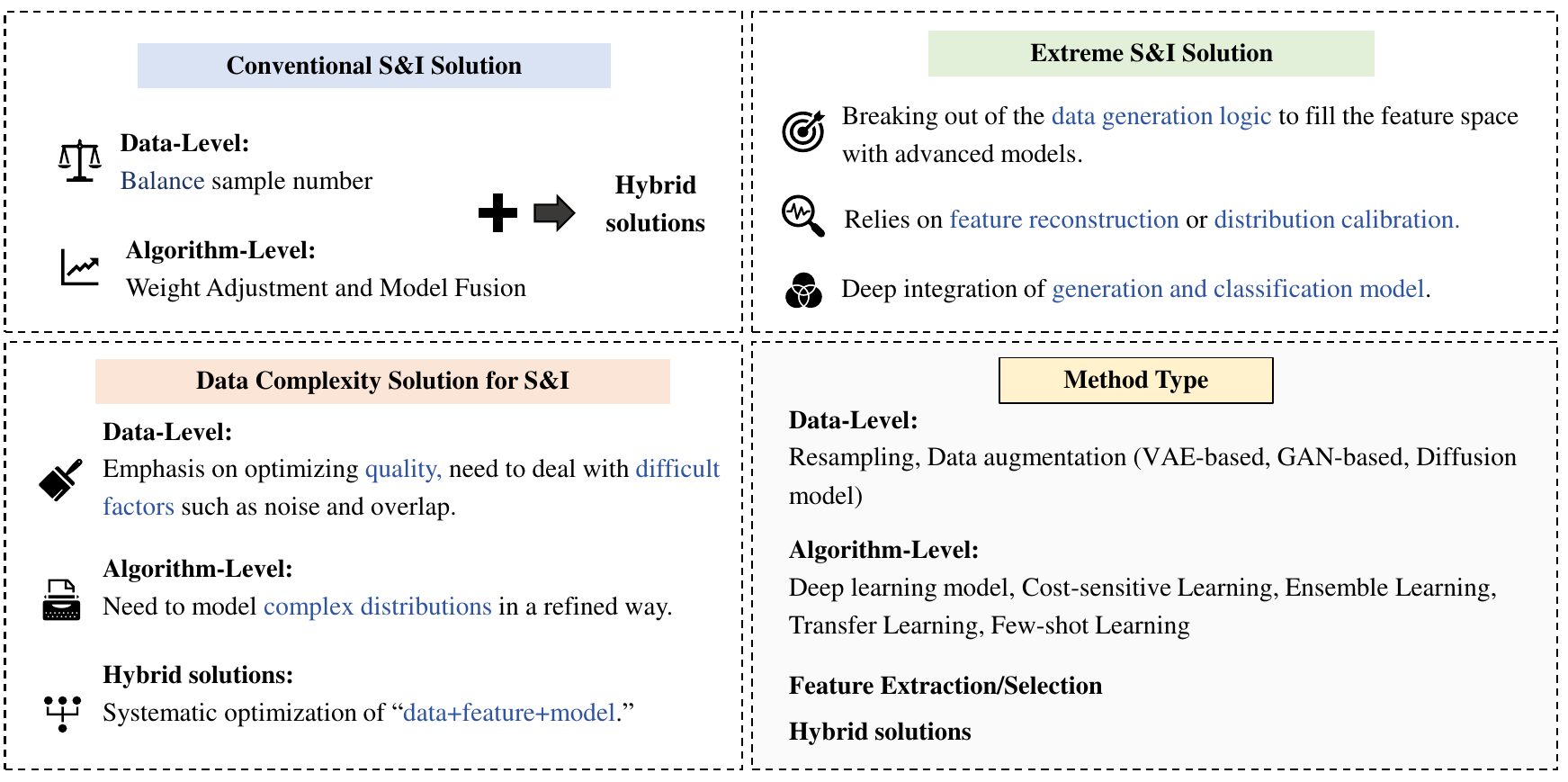}
	\caption{Overview of Solutions Categories and Method Emphasis in S\&I Problem.}
	\label{fig_5}
\end{figure*}

\subsection{Conventional S\&I Solutions}
This section method focuses on addressing small-sample and imbalance issues, with less emphasis on analyzing inter-class and intra-class difficulty factors. However, data augmentation remains an effective strategy for S\&I datasets, as it enhances diversity and generates high-quality samples. Therefore, these methods do not imply an inability to handle complex datasets. The key lies in how they are applied.

\textit{1) Data-level solutions}

Table V summarizes data-level solutions. These methods balance class distributions by modifying or augmenting data to adjust imbalance ratios while maintaining majority class recognition performance \cite{ref74}.

\begin{table}
	\begin{center}
		\caption{Conventional S\&I Solutions on Data-level.}
		\label{tab5}
		\begin{tabular}{lll}
			\hline
			\textbf{Type}&	\textbf{Base model}&	\textbf{Methods or References}\\
			\hline
			Resampling&	Synthetic Oversampling&	SMOTE \cite{ref75}\\ &&Borderline-SMOTE \cite{ref76}\\ 
			&&ADASYN \cite{ref77}\\ 
			&&WK-SMOTE \cite{ref78} \\
			&&MC-SMOTE \cite{ref79}\\
			&Combined Approaches&	SUNDO \cite{ref80}\\
			&& WRO \cite{ref81}\\
			\hline
			\multirow{2}{*}{\makecell[l]{Data \\augmentation}} &	Traditional method&	\cite{ref82}\\
			&VAE-based&	\cite{ref83}\\
			&&WM-CVAE \cite{ref84}\\ 
			&&TL-VAE \cite{ref85}\\
			&GAN-based&	DAC \cite{ref86}\\ 
			&&\cite{ref87}\\
			&&ACGAN-SN \cite{ref88}\\
			&&SCA-GAN \cite{ref89}\\
			&&AGMAN \cite{ref90}\\ 
			&&ConvGeN \cite{ref91}\\
			&Diffusion Model&	DDPM \cite{ref94}\\
			\hline
		\end{tabular}
	\end{center} 
\end{table}

\textit{Resampling:} Undersampling is less commonly applied to S\&I datasets because it balances data by reducing majority-class samples. When data is limited, this can lead to the loss of valuable information and negatively impact classification performance. In practice, oversampling is applied only to the training set, rather than the entire dataset, to avoid overly optimistic performance estimates \cite{ref35}.

The Synthetic Minority Over-Sampling Technique (SMOTE) \cite{ref75} is a widely adopted resampling technique. It generates new minority samples by interpolating in the feature space to balance the dataset. However, it has limitations, including neglecting class distribution complexity and producing redundant or noisy samples. To address these issues, several extensions have been developed. Borderline-SMOTE \cite{ref76} enhances classification by generating samples near decision boundaries, improving robustness in critical regions. Adaptive Synthetic Sampling (ADASYN) \cite{ref77} further refines sample generation by prioritizing hard-to-classify instances, better aligning with the data distribution. The Weighted Kernel-based SMOTE (WK-SMOTE) algorithm \cite{ref78} improves SMOTE by performing oversampling in the feature space of a Support Vector Machine (SVM). This approach effectively addresses SMOTE limitations in nonlinear problems. Minority Clustering SMOTE (MC-SMOTE) \cite{ref79} applies K-means clustering to minority-class samples and generates new samples between neighboring clusters. It balances data distribution and enhances classification performance. However, these do not explicitly address uncertainty issues.

Beyond SMOTE-based techniques, Similarity-based UnderSampling and Normal Distribution-based Oversampling (SUNDO) \cite{ref80} reduce information loss by synthesizing samples near minority instances and removing redundant ones based on similarity. SUNDO is limited to binary classification and is particularly effective for highly imbalanced datasets but struggles with more complex and imbalanced scenarios. The windowed Regression Over-Sampling (WRO) method \cite{ref81} generates virtual samples in minority classes. In contrast to traditional SMOTE, WRO computes regression coefficients in a localized region, considers additive and multiplicative effects, and smoothes the process with a Savitzky-Golay filter to reduce noise. In summary, resampling techniques have long been fundamental methods for addressing S\&I issues. Researchers have explored many approaches, each characterized by distinct features, although not necessarily applicable to every problem.

\textit{Data augmentation} enhances model generalization by transforming and expanding the original dataset to generate additional samples. These methods are particularly effective in addressing class imbalance and small-sample challenges, improving model performance and robustness. Basic approaches include image rotation, scaling, and translation. Fu et al. \cite{ref82} first preprocessed the images by applying rotations, flips, translations, and random erasures. Then, they employed the Slab Defect Generation (SDG) model—an encoder-decoder structure with a discriminator—to perform high-dimensional image fitting and generation, which is particularly suitable for small sample datasets. With technological advancements, deep generative models such as VAEs, GANs, and diffusion models have demonstrated superior capabilities in synthetic data generation.

Wan et al. \cite{ref83} proposed a VAE-based synthetic data generation method that learns the data distribution to generate new samples, thereby balancing the dataset. Compared to traditional SMOTE, VAE-generated images exhibit greater clarity and sharpness in high-dimensional image processing. The Weighted Modified Conditional Variational Autoencoder (WM-CVAE) \cite{ref84} solves KL divergence vanishing using an adaptive loss function. It also applies the Kernel Mean Matching (KMM) algorithm to weight generated samples, reducing dissimilar sample effects. WM-CVAE shows strong robustness across different imbalance ratios, improving fault diagnosis accuracy. Transfer Learning-based VAE (TL-VAE) \cite{ref85} pre-trains the VAE on normal data and fine-tunes it with limited fault samples to generate high-quality fault data for balancing sample distributions. However, the coverage capability of TL-VAE requires further optimization to enhance applicability across broader operating conditions.

GANs can efficiently learn complex data to generate images and perform well in S\&I tasks. Data Augmentation Classifier (DAC) \cite{ref86} innovatively integrates supervised learning with GAN-based data generation, creating an end-to-end model. The framework uses data filtering and purification to mitigate GAN randomness. Its enhanced version, MDAC, uses multiple generators for one-to-one minority sample generation, overcoming single-generator limitations in complex imbalanced scenarios. Nazari et al. \cite{ref87} applied Conditional GAN (CGAN) for targeted upsampling in cybersecurity, effectively handling extreme imbalance. Follow-up studies show a trend towards technology convergence. The Auxiliary Classifier Generative Adversarial Network with Spectral Normalization (ACGAN-SN) \cite{ref88} improves sample fidelity by incorporating spectral normalization, generating fault data nearly indistinguishable from actual samples. SCA-CGAN \cite{ref89} integrates side-channel analysis (SCA) with CGAN to generate specific energy traces, effectively tackling small-sample issues in detection tasks. Further advancing GAN architectures, Auxiliary Generative Mutual Adversarial Networks (AGMAN) \cite{ref90} combine autoencoders (AE) with dual-discriminator GANs (D2GAN). It leverages an auxiliary generator to enhance sample quality while employing adversarial training for improved stability. Convex-space Generative Network (ConvGeN) \cite{ref91} redefines GAN frameworks. It introduces a convex-space generation strategy that allows joint optimization with the discriminator, significantly improving generalization. These approaches improve minority-class recognition while maintaining majority-class performance. GAN-based methods have evolved from simple data augmentation to more refined feature-space optimization, reflecting a broader trend toward enhancing sample quality and model robustness.

In recent years, diffusion models \cite{ref92, ref93} have shown strong performance in data augmentation and generation tasks. It adds noise to the data and then denoises it to learn the underlying distribution. This approach enables the generation of high-quality and realistic synthetic samples. A novel data augmentation method based on the Denoising Diffusion Probabilistic Model (DDPM) addresses data imbalance in industrial fault diagnosis \cite{ref94}. DDPM generates samples through physical simulation, unlike GANs that rely on adversarial training. This avoids instability during training and improves sample authenticity and diversity. 

\textit{2) Algorithm-level solutions}

Algorithm-level optimization improves model design to mitigate data imbalance, ensuring robustness and efficiency with limited data. Table VI summarizes related solutions. The DSLWCN-VAFL framework \cite{ref95} integrates deep separable Laplace wavelet convolution (DSLWC) and variable asymmetric focal loss(VAFL), offering a lightweight network solution. The DSLWC layer extracts time-frequency features. The VAFL dynamically adjusts the contributions of different class samples to mitigate the model bias toward majority classes. DSLWCN-VAFL demonstrates outstanding diagnostic performance and noise resistance while maintaining low computational resource consumption. Bai et al. \cite{ref96} proposed an improved Vision Transformer method incorporating spatial pooling. It effectively reduces the number of model parameters and enhances the generalization capability. Wang et al. \cite{ref97} developed an intelligent fault diagnosis method based on the Stacked Capsule Autoencoder (SCAE) to address S\&I challenges in wind turbine fault diagnosis. It optimizes spectral feature extraction and fault classification by leveraging unsupervised learning and pose-aware adjustments.

\begin{table}
	\begin{center}
		\caption{Conventional S\&I Solutions on Algorithm-level.}
		\label{tab6}
		\begin{tabular}{ll}
			\hline
			\textbf{Type}& \textbf{Methods or References}\\
			\hline
			Deep learning model&	DSLWCN-VAFL \cite{ref95}, \cite{ref96}, \cite{ref97}\\
			Cost-sensitive learning&	\cite{ref98}\\
			Ensemble learning&	\cite{ref25}\\
			Transfer learning&	\makecell[l]{Rare-Transfer \cite{ref102}, CIATL\cite{ref103},\\ \cite{ref104}, G-TELM \cite{ref105}}\\
			Few-shot learning&	PTAFedIF \cite{ref106}, TRN \cite{ref107}\\
			\hline
		\end{tabular}
	\end{center} 
\end{table}

\textit{Cost-sensitive learning} improves model focus on minority classes through a differentiated weighting mechanism. The Class-wise Difficulty-Balanced Loss (CDB Loss) \cite{ref98} introduces an innovative dynamic inter-class difficulty assessment strategy. This method quantifies the learning difficulty of each class during training, which breaks away from the traditional paradigm of assigning weights based on sample numbers. It adaptively adjusts sample weights to strengthen the representation learning of difficult classes, where it outperforms conventional loss functions.

\textit{Ensemble learning} \cite{ref99} enhances model performance and robustness by aggregating predictions from multiple base learners. It is particularly effective in handling complex or imbalanced datasets, as it reduces overfitting and enhances classification accuracy. Sun et al. \cite{ref25} proposed a method that divides an imbalanced dataset into several balanced subsets and trains multiple classifiers on these subsets. The results are combined using a specific ensemble rule, such as MaxDistance. It does not alter the original class distribution, thus avoiding information loss and overfitting while demonstrating improved classification performance and generalization. It outperforms traditional Bagging \cite{ref100} and Boosting \cite{ref101} ensemble techniques on highly imbalanced datasets.

\textit{Transfer learning} enhances minority class classification in small datasets by fine-tuning pre-trained models on large datasets. Rare-Transfer \cite{ref102} leverages auxiliary data and a label-dependent weight update mechanism to address class imbalance and insufficient training samples. Class-Imbalance Adversarial Transfer Learning (CIATL) \cite{ref103} is a cross-domain fault diagnosis framework that integrates class imbalance learning and dual-layer adversarial training. It improves minority class classification by extracting domain-invariant and class-separating features from imbalanced source data. Sima et al. \cite{ref104} developed an edge-cloud collaborative detection method that enhances CNN-based fault detection on S\&I datasets. This approach combines transfer learning and federated learning, integrating local edge node fine-tuning with global cloud aggregation. It maintains high performance over varying degrees of S\&I. Generalized Transfer Extreme Learning Machine (G-TELM) \cite{ref105} is a novel domain adaptation method for unsupervised cross-domain fault diagnosis. G-TELM minimizes the differences in source-target domain distributions, employs cost-sensitive classification to reduce misclassification losses, and incorporates L2 regularization to enhance transferability. These approaches significantly improve performance on S\&I datasets.

\textit{Few-shot learning} enhances model performance in data-scarce environments through efficient feature extraction and knowledge transfer. Pretraining Adaptive Federated Learning Incipient Fault Identification (PTAFedIF) \cite{ref106} uses pretraining to improve few-shot learning. It adopts a federated learning framework to protect privacy and enable cross-regional collaboration. Adaptive weighting factors are introduced to enhance model generalization. Transfer Relation Network (TRN) \cite{ref107} combines few-shot and transfer learning for rotating machinery fault diagnosis. The TRN uses a relational network and multi-kernel maximum mean discrepancy (MK-MMD) for efficient feature extraction and domain adaptation, enhancing fault diagnosis accuracy and transfer performance.

\textit{3) Hybrid solutions}

Some have leveraged hybrid methods that combine algorithms, resampling techniques, and deep learning models to address S\&I challenges. Martin-Diaz et al. \cite{ref108} proposed a classification approach integrating AdaBoost with optimized sampling. The method significantly improves induction motor fault detection while maintaining robustness to data scale and imbalance ratios. Pérez-Ortiz et al. \cite{ref109} combined synthetic data with semi-supervised learning (SSL), achieving superior classification performance in high-dimensional S\&I scenarios. Wang et al. \cite{ref110} proposed a bearing fault diagnosis method that combines an improved SMOTE and CNN-Attention Mechanism (CNN-AM). A Gaussian mixture model-optimized SMOTE enhances minority class diversity, while an SVM-based technique refines the data generation boundaries. CNN-AM efficiently extracts fault features for high-precision classification. This method outperforms traditional GAN in terms of processing speed.

Some researchers have focused on integrating data augmentation, deep learning, and generative models to improve performance and address challenges like S\&I. Kim et al. \cite{ref111} selected wafer maps with clear defect patterns and augmented the data using zero-padding, rotation, and flipping. Training a CNN classifier on the enhanced dataset significantly improved classification performance. DeepSMOTE \cite{ref29} embeds SMOTE into Deep Convolutional GAN (DCGAN), generating samples in the low-dimensional feature space and optimizing the loss function to improve imbalance handling. This approach outperforms traditional pixel-level methods and GAN-based baselines. DCNN-VC \cite{ref112} combines deep CNN (DCNN) with voting classification. This method uses a sliding window technique for time-domain data augmentation of vibration signals while preserving key features. DCNN extracts deep features, and ensemble voting enhances classification reliability. Balanced GAN and Gradient Penalty (BAGAN-GP) \cite{ref113} combines balanced GANs with transfer learning. This framework improves training stability through gradient penalty and addresses the S\&I problem in spot welding defect classification. Yao et al. \cite{ref114} introduced a skin lesion classification model. It integrates an enhanced DCNN, RandAugment, a multi-weight reweighted loss function (MWNL), and a cumulative learning strategy (CLS), improving accuracy while enabling low-cost screening. Qin et al. \cite{ref115} designed a hybrid framework. It combines a Wasserstein distance-based deep convolutional GAN with gradient penalty (WDCGAN-GP) and a CNN with Coordinate Attention (CNN-CA). The self-attention mechanism and spectral normalization improve the quality of the generated bearing fault samples. The coordinate attention network enhances diagnostic accuracy.

Some have focused on combining feature extraction or selection, and data balancing techniques with deep learning to improve classification accuracy and robustness in various fields. Xu et al. \cite{ref116} proposed a fault diagnosis framework for rotating machinery that integrates multi-domain feature extraction, feature selection, and cost-sensitive learning. Wang et al. \cite{ref117} addressed the S\&I issue in fMRI data by integrating Independent Component Analysis (ICA) with deep learning. ICA reduced redundancy, SMOTE balanced the dataset, and a Convolution-GRU network extracted temporal features, outperforming traditional methods such as SVM. SASYNO-RF-RSKNN \cite{ref118} is a hybrid model for traffic event detection. It uses Self-Adaptive Synthetic Over-Sampling Technique (SASYNO) for data balancing, random forest (RF) for feature selection, and RSKNN for classification. PGC-FSDTF \cite{ref119} integrates data augmentation, traditional features (PHOG, HSV histograms), and deep features (DenseNet201, InceptionResNetV2). It combines feature selection and multiple classifiers, improving pedestrian gender classification accuracy and robustness. Zhang et al. \cite{ref120} developed an integrated end-to-end model. It combines VAE, Wasserstein GAN with Gradient Penalty (WGAN-GP), and ensemble classifiers for data augmentation, pseudo-label utilization, and fault detection. Kuo et al. \cite{ref121} validated the superiority of the SMOTE+RF combination for autonomous driving collision prediction. They also proposed an optimized ``balance first, then feature selection'' process.

\textit{Summary:} While conventional methods can partially mitigate issues related to small sample sizes and data imbalance, substantial improvements require a detailed analysis of both data characteristics and algorithm focus. López et al. \cite{ref122} evaluated preprocessing methods and cost-sensitive learning on 66 imbalanced datasets. The results showed no significant difference between these approaches in improving classification performance. Both methods effectively addressed imbalanced data challenges, but their effectiveness varied with the classification algorithm used. In some cases, a hybrid approach combining both methods outperformed individual techniques, although its performance was highly dependent on dataset characteristics and algorithm choice.

\subsection{Data Complexity Solutions for S\&I} 
Class overlap and dataset shift are key factors that contribute to the decline in classification performance, particularly in imbalanced datasets \cite{ref122}. Figure 7 shows these complex factors between the two classes. In multi-class problems, such complexity may appear locally between specific class pairs or globally across multiple classes \cite{ref62}. Dudjak et al. \cite{ref38} identified noise as the most detrimental factor affecting classification performance, followed by class overlap and imbalance. Seiffert et al. \cite{ref123} found that classification algorithms are more sensitive to noise than to class imbalance. Additionally, Gupta et al. \cite{ref124} reported that class overlap is a primary cause of label noise, a finding echoed by other studies \cite{ref56, ref125}. Complex dataset characteristics limit the effectiveness of traditional methods. Therefore, more sophisticated approaches are needed. Existing methods can be categorized into four levels: data-level, feature-level, algorithm-level, and hybrid solutions.
\begin{figure}[!t]
	\centering
	\includegraphics[width=\linewidth]{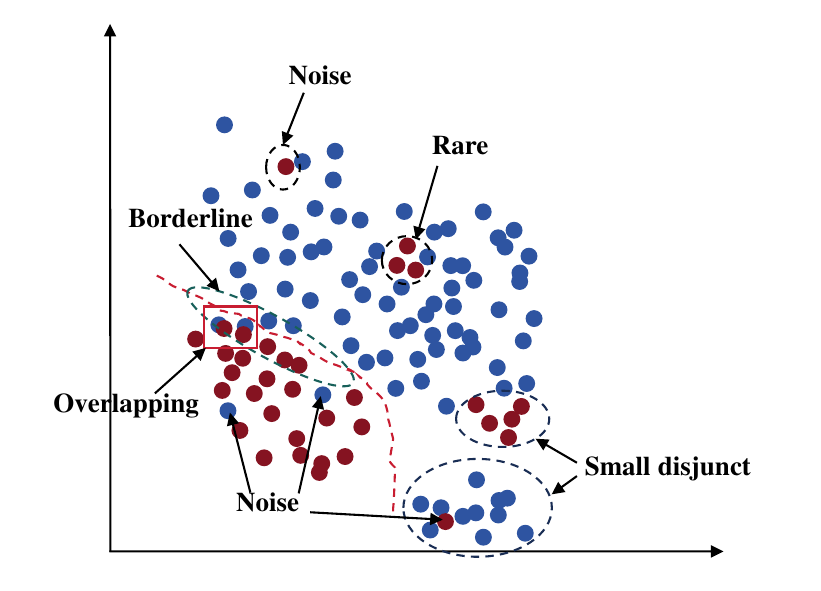}
	\caption{Example of a binary dataset with complex factors.}
	\label{fig_6}
\end{figure}

\textit{1) Data-level solutions}

\textit{Resampling:} Oversampling techniques offer potential advantages in addressing noisy data and class overlap \cite{ref17}. Most resampling methods focus on feature-based strategies when dealing with S\&I data that presents challenging factors. Simple undersampling alone cannot fundamentally resolve these issues, but selectively removing instances from overlapping regions can enhance classification performance \cite{ref126}. García et al. \cite{ref127} demonstrated that oversampling and hybrid strategies are more effective than undersampling in increasing the proportion of safe minority-class samples and reducing unsafe ones. As shown in Table VII, we aggregate and average the experimental results from studies \cite{ref65, ref66, ref128, ref129, ref130} to provide an overview of resampling methods for addressing S\&I problems with complex factors.
\begin{table}
	\begin{center}
		\caption{Resampling Methods for Complex S\&I Datasets.}
		\label{tab7}
		\begin{tabular}{llccc}
			\hline
			\textbf{Type}&	\textbf{Method}&	\textbf{G-mean}&	\textbf{F1-score}&	\textbf{AUC}\\
			\hline 
			Baseline&	No-sampling&	57.18&	48.32&	70.22\\
			\hline
			Filtering-Based &	SMOTE-Tomek&	53.30&	46.50&	75.60\\
			&SMOTE-ENN&	61.41&	45.795&	73.83\\
			&SMOTE-IPF&	69.83&	43.3&	71.58\\
			&SMOTE-WENN& 70.83&	46.38&	72.51\\
			\hline
			Feature-Based&	FW-SMOTE&	80.75&	-&	86.40\\
			\hline
			\makecell[l]{Sample\\Concatenation}&	Re-SC&	77.60&	60.10&	81.00\\
			\hline
			\makecell[l]{Clustering + \\Weighting} &	A-SUWO&	67.96&	57.36&	73.93\\
			&IA-SUWO&	79.44&	71.67&	80.12\\
			&MWMOTE&	74.50&	62.36&	78.60\\
			&NI-MWMOTE&	75.21&	63.23&	77.41\\
			&IMWMOTE&	-&	-&	-\\
			\hline
			Cleaning-Based&	CCR&	-&	-&	-\\
			&RB-CCR&	-&	-&	-\\
			\hline
			Radial-Based&	RBO&	78.8&	-&	83.95\\
			\hline
		\end{tabular}
	\end{center} 
	\footnotesize \textit{Data not retrieved are indicated by “-” and data are percentage values (\%).}
\end{table}

Building on SMOTE, various methods have been developed to handle dataset complexity by improving sample generation and filtering mechanisms. SMOTE-Tomek \cite{ref131} combines oversampling with Tomek link-based filtering to refine class clusters and increase minority class representation. However, excessive filtering may lead to information loss. SMOTE-ENN \cite{ref131} applies the Edited Nearest Neighbor (ENN) rule to remove noisy samples, achieving higher classification accuracy, particularly in high-noise scenarios. SMOTE-IPF \cite{ref132} further refines this approach using an iterative partition filter (IPF) to process noise and boundary samples. SMOTE Weighted ENN (SMOTE-WENN) \cite{ref66} optimizes SMOTE-ENN and SMOTE-IPF by incorporating a weighted distance function and k-nearest neighbor rules. This enhances sample generation and cleaning, leading to superior classification performance. Feature-weighted SMOTE (FW-SMOTE) \cite{ref128} improves synthetic sample generation by integrating feature weighting and using a weighted Minkowski distance, effectively mitigating feature redundancy and noise. Table VII shows its strong performance. The Resampling Algorithm Based on Sample Concatenation (Re-SC) \cite{ref129} reduces class overlap by concatenating samples with the same label into a new feature space, enhancing minority class classification. EnRe-SC further enhances this approach by integrating ensemble learning, achieving better generalization. Experimental results confirm that Re-SC and EnRe-SC outperform traditional SMOTE methods across multiple evaluation metrics.

The following are clustering-based weighted oversampling methods. Adaptive Semi-unsupervised Weighted Oversampling (A-SUWO) \cite{ref133} employs semi-unsupervised clustering to identify minority instances near decision boundaries and adjusts oversampling intensity based on classification complexity. The Improved A-SUWO (IA-SUWO) \cite{ref65} incorporates Least-Squares SVM (LS-SVM) values and the Improved Majority Weighted Minority Oversampling (IMWMO) to refine the weighting of minority class instances. It also uses a kINN approach to synthesize new instances, improving boundary instance generation and increasing sample quality and diversity. Table VII shows strong performance. Majority Weighted Minority Oversampling Technique (MWMOTE) \cite{ref134} identifies hard-to-learn minority class samples and assigns weights based on their Euclidean distance to majority class samples. Noise-Immunity MWMOTE (NI-MWMOTE) \cite{ref130} enhances MWMOTE by incorporating adaptive noise handling, agglomerative hierarchical clustering, and adaptive sub-cluster sizing. It is particularly effective for highly noisy and imbalanced data, significantly addressing inter-class and intra-class imbalances. Improved MWMOTE (IMWMOTE) \cite{ref20} uses unsupervised clustering and k-NN denoising to segment samples and enhance data quality. By determining optimal sampling sizes for subsets, IMWMOTE effectively tackles heterogeneous S\&I fault modeling, improving model interpretability. The Expectation Maximization-based Local-Weighted Minority Oversampling Technique (EM-LWMOTE) \cite{ref135}  identifies hard-to-learn, information-rich minority fault samples in multi-class fault diagnosis. It increases the representativeness of synthetic samples by generating them within minority class clusters and boosts classification performance through dynamic oversampling and adaptive weighting.

Several approaches address the difficult factors in S\&I by combining cleaning and resampling. Combined Cleaning and Resampling (CCR) \cite{ref136} improves minority class detection by removing majority class samples near minority instances and generating synthetic samples around unsafe regions. It is effective for numerical data and binary classification. Multi-class CCR (MC-CCR) \cite{ref137} extends CCR by optimizing class boundaries, reducing class overlap, and proportionally generating synthetic samples within cleaned regions. It performs well in handling label noise, particularly near boundary and outlier points. Radial-based CCR (RB-CCR) \cite{ref138} introduces Radial Basis Functions (RBF) to partition the data space based on class potential, guiding synthetic sample generation. However, its parameter selection significantly affects performance. Radial-based oversampling (RBO) \cite{ref139} leverages RBF to identify suitable regions for generating minority class samples while minimizing class overlap. It is effective in handling label and feature noise. Multi-class RBO (MC-RBO) \cite{ref140} further enhances multi-class imbalance handling by computing class potential to generate artificial samples in regions with high classification uncertainty.

Multi-class classification adds complexity to S\&I data. Sáez et al. \cite{ref62} analyzed oversampling techniques for multi-class imbalanced datasets. They found that traditional uniform strategies are ineffective in complex scenarios because they treat all classes and instance types equally. They proposed a selective oversampling approach that identifies different instance types, such as safe, borderline, rare, and outlier samples, and applies targeted oversampling. This approach exhibits robustness across multiple datasets with specific configurations. The Mahalanobis Distance-based Over-sampling (MDO) technique \cite{ref63} addresses the majority class by preserving the covariance structure of the minority class. It generates synthetic samples along probability iso-contours, thereby reducing class overlap. The Adaptive Mahalanobis Distance-based Over-sampling (AMDO) \cite{ref141} improves MDO by capturing minority class covariance and generating synthetic samples along probability contours. It uses Generalized Singular Value Decomposition for mixed-type datasets and optimizes sample synthesis through partial balancing resampling, boosting classifier accuracy, and performance.

We analyze typical methods for addressing multi-class problems, which are presented in the experimental results from \cite{ref29}. The experiments use the average results from five datasets—MNIST \cite{ref142}, FMNIST \cite{ref143}, CIFAR-10 \cite{ref144}, SVHN \cite{ref145}, and CelebA \cite{ref146}. All resampling methods employ the same ResNet-18 \cite{ref147} classifier. Table VIII shows that methods outperform baseline SMOTE across all metrics. AMDO improves Recall and F1-score by focusing on difficult-to-learn instances. MC-RBO achieves the highest performance in all metrics, demonstrating the effectiveness of radial-based oversampling in multi-class S\&I datasets. 

Besides resampling, other solutions addressing S\&I data complexity are summarized in Table IX.

\begin{table}
	\begin{center}
		\caption{Resampling Methods for Multi-Class S\&I Complex Datasets.}
		\label{tab8}
		\begin{tabular}{llccc}
			\hline
			\textbf{Type}&	\textbf{Method}&	\textbf{Recall}&	\textbf{G-Mean}&	\textbf{F1-score}\\
			\hline
			Baseline&	SMOTE&	60.574&	68.012&	56.930\\
			Distance-based&	AMDO&	63.240&	71.688&	60.920\\
			Cleaning-based&	MC-CCR&	66.568&	74.482&	65.778\\
			Radial-based&	MC-RBO&	68.854&	78.196&	68.876\\
			\hline
		\end{tabular}
	\end{center} 
	\hspace{0.6cm}
	\footnotesize \textit{Data are percentage values (\%).}
\end{table}

\textit{Data Augmentation:} The generation model of this part generates effective samples to handle S\&I datasets with complex factors through various mechanisms and loss functions. The Minority Oversampling GAN (MoGAN) \cite{ref148} uses a generator to create minority class samples and a discriminator to detect faults. It employs a feature-matching loss function to generate samples in low-density regions, thereby improving detection performance. Full Attention Wasserstein GAN with Gradient Normalization (FAWGAN-GN) \cite{ref149} improves training stability with Gradient Normalization and leverages Full Attention to generate high-quality samples. Parallel Classification Wasserstein GAN with Gradient Penalty (PCWGAN-GP) \cite{ref150} assigns independent generators, discriminators, and classifiers to each failure class. It optimizes sample generation using Pearson and separability loss functions, achieving high accuracy and robustness across different imbalance ratios. Categorical Feature GAN (CFGAN) \cite{ref151} generates high-quality synthetic samples to balance the dataset by learning classification features from limited faulty samples and interpolating with noise in the feature space.

\begin{table}
	\begin{center}
		\caption{Other Data Complexity Solutions for S\&I besides Resampling.}
		\label{tab9}
		\begin{tabular}{ll}
			\hline
			\textbf{Type}&	\textbf{Methods or References}\\
			\hline
			Data Augmentation&	MoGAN \cite{ref148}\\
			&FAWGAN-GN \cite{ref149}\\
			&PCWGAN-GP \cite{ref150}\\
			&CFGAN \cite{ref151}\\
			\hline
			Feature-level	& S2N and FAST \cite{ref152}\\
			&AMF \cite{ref154}\\
			&DBFS \cite{ref155}\\
			&UIG-CFGVM \cite{ref156}\\
			&MOSNS and MOSS \cite{ref157}\\
			\hline
			Algorithm-level	&Soft-Hybrid \cite{ref159}\\
			&Meta-learning \cite{ref160}\\
			&Class-Balanced loss \cite{ref161}\\
			&RBBag \cite{ref162}\\
			&RBBag+RSM and MRBBag \cite{ref163}\\
			\hline
			Hybrid solutions&	WMODA \cite{ref164}\\
			&PCA-SMOTE-SVM \cite{ref165}\\
                & Hartono et al. \cite{ref166}\\
			&Qin et al. \cite{ref167}\\
			&Zhang et al. \cite{ref168}\\
			&Wei et al.: \cite{ref20}, \cite{ref169, ref170, ref171}\\
			\hline
		\end{tabular}
	\end{center} 
\end{table}

\textit{2) Feature-level solutions}

Feature selection is a crucial method for addressing the complexity of S\&I problems. It reduces dimensionality and enhances classifier performance by selecting the most discriminative features. Wasikowski and Chen \cite{ref152} found that Signal-to-Noise Correlation (S2N) and Fast Sliding Threshold Feature Evaluation (FAST) \cite{ref153} are effective feature selection methods, especially for selecting a small number of features. The AUC Margin Fraction (AMF) \cite{ref154} is based on maximizing the AUC margin through Boosting learning. It evaluates feature importance using weighted linear combinations, thereby optimizing the feature selection process. Density-Based Feature Selection (DBFS) \cite{ref155} assesses feature contribution through probability density estimation, which aims to maximize class separation and minimize overlap. The Union Information Gini Cost-sensitive Feature Selection General Vector Machine (UIG-CFGVM) \cite{ref156} combines the speed and scalability of Filter algorithms with the interactivity and feature dependency of Wrapper algorithms, improving performance in high-dimensional S\&I data. Minimizing Overlapping Selection under No-Sampling (MOSNS) and Minimizing Overlapping Selection under SMOTE (MOSS) \cite{ref157} use sparse regularization techniques, such as elastic net penalties, to reduce feature overlap between classes, enhancing performance in imbalanced learning.

\textit{3) Algorithm-level solutions}

When class overlap is high, collecting more training data is unlikely to improve classification accuracy \cite{ref158}. Therefore, an effective recognition algorithm also plays a crucial role. The Soft-Hybrid algorithm \cite{ref159} enhances classification in highly imbalanced datasets. It segments the data into non-overlapping, boundary, and overlapping regions and applies tailored classification strategies to improve accuracy. Ren et al. \cite{ref160} proposed a meta-learning algorithm that enhances deep learning model robustness by automatically adjusting training sample weights via meta-gradient descent. Cui et al. \cite{ref161} developed Class-Balanced Loss, which balances the loss by inversely weighting each class's effective sample count. This non-parametric approach eliminates distributional assumptions. The key idea is to quantify data overlap to estimate the effective sample size. As a result, it is widely applicable to existing models and loss functions.

Ensemble learning methods have been extensively explored to enhance the robustness and adaptability of classifiers. Roughly Balanced Bagging (RBBag) \cite{ref162} handles unsafe minority class samples, such as boundary, rare, and outlier instances, while maintaining robustness to base classifier selection. Lango and Stefanowski \cite{ref163} enhanced RBBag’s performance on high-dimensional datasets and improved classifier diversity by integrating it with the Random Subspace Method (RBBag+RSM). They introduced Multi-class Roughly Balanced Bagging (MRBBag) to address the multi-class imbalance. This method adjusts the proportion of each class in bootstrap samples and significantly improves classification performance on multi-class datasets.

\textit{4) Hybrid solutions}

This section focuses on addressing complex S\&I problems by integrating data balancing, feature extraction, and classifier improvement. WMODA \cite{ref164} uses Weighted Minority Oversampling (WMO) to balance the data, followed by an enhanced Deep Auto-Encoder (DA) for feature extraction and a C4.5 decision tree for classification. While effective for fault detection, WMODA is limited to binary classification and is sensitive to data imbalance. The PCA-SMOTE-SVM model \cite{ref165} combines PCA, SMOTE, and SVM to improve diagnostic accuracy. Hartono et al. \cite{ref166} proposed a new method that combines feature selection with a Hybrid Approach Redefinition (HAR). This method optimizes the ensemble learning algorithm and preprocessing steps in HAR by introducing MOSS. This effectively reduces class overlap and thus improves classifier performance. Qin et al. \cite{ref167} proposed a weakly supervised oversampling framework. It employs Cost-Sensitive Neighborhood Component Analysis (CS-NCA) for dimensionality reduction. The framework also integrates boosting-based ensemble learning to reduce imbalance and enhance classification performance. Zhang et al. \cite{ref168} presented an ensemble resampling-based transfer AdaBoost algorithm for credit S\&I classification. It employs neighborhood-based filtering and bagging resampling to generate balanced datasets. Additionally, it utilizes a weight-adaptive TrAdaBoost algorithm to enhance classification performance.

Wei et al. proposed a series of innovative hybrid methods aimed at addressing challenges such as small sample sizes, class imbalance, and noise interference in rolling bearing fault diagnosis. In 2020, they combined Cluster-MWMOTE, Moth-Flame Optimization (MFO), and LS-SVM to balance minority class samples and optimize LS-SVM hyperparameters \cite{ref169}, improving efficiency and reducing overfitting. In 2021, the team introduced a new diagnostic method based on Sample-characteristic Oversampling Technique (SCOTE) and Multi-Class LS-SVM \cite{ref170}. This method converts multi-class imbalance into binary imbalance. It uses KNN to filter noise and evaluates sample importance with LS-SVM, thereby enhancing model robustness and accuracy. In 2024, they further explored noise and data scarcity challenges with a fault diagnosis method based on NI-MWMOTE \cite{ref171}. It optimizes hyperparameters and noise processing for real-world monitoring. They also proposed combining IMWMOTE with LS-SVM to improve performance on imbalanced data and mixed class distributions \cite{ref20}. These methods demonstrate Wei et al.'s continuous advancements in improving fault diagnosis through hybrid techniques that address key challenges in real-world datasets.

\textit{Summary:} These methods optimize the expressive power of the model by mining deep data features, making them effective for high-dimensional or complex datasets. However, feature-based methods are inevitably affected by the quality and dimensionality of the input features. The generalization ability of the model can also be affected by the degree of class imbalance. The degree of difficulty factors can also hinder the effectiveness of learning. Therefore, these techniques are typically adapted to dataset-specific challenges and optimized through iterative experimentation to achieve optimal results.

\subsection{Extreme S\&I Solutions}

Dealing with extremely S\&I datasets presents significant challenges. Effective approaches often require the integration of domain knowledge, in addition to data augmentation and sampling techniques. They may also involve using more sophisticated model architectures and training strategies. In some cases, redefining and reframing the problem may be necessary. Thabtah et al. \cite{ref74} found that classifiers achieve lower error rates on highly imbalanced datasets than on balanced datasets with the same number of samples. This occurs because the majority class dominates the learning process, while the minority class is largely overlooked. Consequently, numerous methods have been proposed to address the problem of extreme class imbalance.

Sampling WIth the Majority (SWIM) \cite{ref19} generates new samples along equal-density contours using Mahalanobis distance. It relies on the positional relationship between the minority and majority classes rather than the minority class distribution. This makes it particularly effective when minority-class samples are scarce. Few-shot GAN \cite{ref172} integrates GANs with transfer learning. It is pre-trained on data-rich categories and fine-tuned using anchor samples to maintain realism and preserve complex distributions. Additionally, it uses large-margin learning to prevent feature shifts. This effectively addresses overfitting and improves sample diversity and accuracy in few-shot scenarios. Uncorrelated Cost-sensitive Multiset Learning (UCML) \cite{ref173} reduces sample correlation by randomly partitioning the dataset. An improved version, Deep Metric-based UCML (DM-UCML), integrates GAN and deep metric learning to ensure consistency in subset distributions and handle nonlinear issues, further improving classification accuracy. The Dual-Stream Adaptive Deep Residual Shrinkage Vision Transformer with Interclass–Intraclass Rebalancing Loss (DSADRSViT-IIRL) \cite{ref174} removes redundant signals and extracts key local features. It also applies Interclass–Intraclass Rebalancing Loss (IIRL) and online hard sample mining to improve recognition of minority class samples.

\textit{Long-tail distribution} \cite{ref175} describes an extreme imbalance in datasets. Most data cluster on the left side of the distribution, while a small number of extreme data points are located in the ``long tail'' region on the right side. Recent advancements in addressing long-tailed distributions include several innovative methods. Invariant Feature Learning (IFL) \cite{ref176} focuses on learning attribute-invariant features to address class and attribute-wise imbalances in long-tailed classification. It serves as a strong baseline for generalized tasks. Representation Calibration methods RCAL \cite{ref177} tackles noisy labels and long-tailed data by calibrating deep representations via contrastive learning and distributional recovery. It recovers underlying distributions and samples additional data points to balance the classifier, significantly improving robustness. Proximate Long-Tail Distribution (ProLT) \cite{ref178} transforms compositional zero-shot learning into a class imbalance problem by estimating attribute priors to adjust posterior probabilities. This enhances performance without additional parameters. The Decoupled Optimization (DO) \cite{ref180} framework balances parameter importance across classes by optimizing different parameter groups for specific class partitions, achieving superior performance. Probabilistic Contrastive (ProCo) \cite{ref181} learning algorithm estimates class distributions and samples contrastive pairs based on the assumption that features follow a von Mises-Fisher (vMF) distribution. It overcomes the large batch size requirement of supervised contrastive learning and achieves strong performance in long-tailed visual recognition. These methods push the state of the art by using diverse strategies and proving effective across various domains and datasets.

\subsection{Experiment}
\begin{table*}
	\begin{center}
		\caption{Experimental Results on the Binary Breast\_cancer Dataset.}
		\label{tab10}
		\begin{tabular}{ l|ccccc| ccccc }
			\hline	
			\textbf{Evaluation}& \textbf{Accuracy} &&&&  &\textbf{AUC}&&&&\\
			\hline								
			Method&	DT&	SVM&	RF&	KNN&	AdaBoost&
			DT&	SVM&	RF&	KNN&	AdaBoost\\
			\hline
			No sampling&	0.9415&	0.9357&	0.9708&	0.9591&	0.9766&
			0.9438&	\textbf{0.9931}&	0.9968&	0.9953&	0.9962\\
			SMOTE&	0.9357&	0.9415&	0.9708&	0.9532&	\textbf{0.9825}&
			0.9357&	0.9415&	0.9708&	0.9532&	0.9825\\
			Borderline\_SMOTE&	0.9123&	0.9123&	0.9649&	0.9357&	0.9591&
			0.9123&	0.9123&	0.9649&	0.9357&0.9591\\				
			ADASYN&	\textbf{0.9649}&	0.9123&	0.9649&	0.9357&	0.9474&
			0.9649&	0.9123&	0.9649&	0.9357&	0.9474\\
			SUNDO&	0.9357&	0.9357&	0.9708&	0.9532&	0.9532&
			0.9357&	0.9357&	0.9708&	0.9532&	0.9532\\
			A\_SUWO&	\textbf{0.9649}&	0.9064&	0.9649&	0.9415&	0.9591&
			\textbf{0.9623}&	0.9893&	0.9979&	0.9853&	0.9951\\
			MWMOTE&	0.9298&	0.9123&	0.9708&	0.9298&	0.9415&
			0.9279&	0.9921&	0.9964&	0.9826&	0.9862\\
			CCR&	0.9474&	\textbf{0.9474}&	\textbf{0.9766}&	0.9474&	0.9591&
			0.9583&	0.9921&	0.9977&	0.9924&	0.9925\\
			SMOTE\_TomekLinks&	0.9298&	\textbf{0.9474}&	0.9708&	0.9649&	0.9591&
			0.9378&	0.9921&	0.9979&	\textbf{0.9966}& 0.9915\\					
			SMOTE-ENN&	0.9357&	0.9357&	0.9708&	\textbf{0.9766}&	0.9708&
			0.9458&	0.9915&	\textbf{0.9987}&	0.9939&	0.9975\\
			SMOTE-IPF&	0.9357&	0.9415&	0.9708&	0.9532&	\textbf{0.9825}&
			0.9425&	0.9921&	0.9975&	0.9934&	\textbf{0.9981}\\
			MDO&	0.9240&	0.9298&	0.9708&	0.9591&	0.9591&
			0.9266&	0.9910&	0.9946&	0.9953&	0.9909\\
			\hline
			\hline
			\textbf{Evaluation}& \textbf{F1-score} &&&&&  \textbf{G-Mean}&&&&\\
			\hline
			Method&	DT&	SVM&	RF&	KNN&	AdaBoost&
			DT&	SVM&	RF&	KNN&	AdaBoost\\
			\hline
			No sampling&	0.9419&	0.9342&	0.9706&	0.9587&	0.9766&
			0.9437&	0.9085&	0.9632&	0.9468&	0.9748\\
			SMOTE& 0.9363&	0.9413&	0.9708&	0.9535&	0.9825&
			0.9421&	0.9334&	0.9702&	0.9563&	0.9828\\
			Borderline\_SMOTE&	0.9131&	0.9136&	0.9650&	0.9364&	0.9594&
			0.9171&	0.9255&	0.9656&	0.9450&	0.9641\\
			ADASYN&	\textbf{0.9650}&	0.9134&	0.9650&	0.9365&	0.9476&
			\textbf{0.9650}&	0.9229&	0.9656&	0.9477&	0.9484\\
			SUNDO&	0.9361&	0.9345&	0.9707&	0.9532&	0.9535&
			0.9391&	0.9130&	0.9668&	0.9496&	0.9563\\
			A\_SUWO&	0.9649&	0.9077&	0.9649&	0.9421&	0.9593&
			0.9623&	0.918&	0.9623&	0.9498&	0.9610\\
			MWMOTE&	0.9300&	0.9134&	0.9708&	0.9307&	0.9420&
			0.9279&	0.9229&	0.9702&	0.9402&	0.9469\\
			CCR&	0.9480&	\textbf{0.9475}&	\textbf{0.9765}&	0.9477&	0.9594&
			0.9574&	\textbf{0.9451}&	0.9714&	0.9516&	0.9641\\
			SMOTE\_TomekLinks&	0.9306&	0.9471&	0.9708&	0.9651&	0.9593&
			0.9373&	0.9379&	0.9702&	0.9688&	0.9610\\
			SMOTE-ENN&	0.9364&	0.9356&	0.9709&	\textbf{0.9767}&	0.9708&
			0.9450&	0.9289&	\textbf{0.9735}&	\textbf{0.9782}&	0.9702\\
			SMOTE-IPF&	0.9363&	0.9413&	0.9708&	0.9535&	\textbf{0.9825}&
			0.9421&	0.9334&	0.9702&	0.9563&	\textbf{0.9828}\\
			MDO&	0.9245&	0.9280&	0.9707&	0.9587&	0.9593	&
			0.9265&	0.8997&	0.9668&	0.9468&	0.9610	\\		
			\hline 
		\end{tabular}
	\end{center}
	\hspace{1cm}
	\footnotesize \textit{Bolded values are the best among different sampling methods for the same classifier with the same evaluation metrics.}
\end{table*}
Different levels of approaches have different applicability and advantages. However, research has shown that resampling methods have long been the most commonly used approach for solving the S\&I problem. In this study, we evaluate several widely used resampling techniques. Specifically, we compare three traditional methods: SMOTE, Borderline-SMOTE, and ADASYN, and seven strategies designed to handle data complexity. The experiments use five classifiers: Decision Tree (DT), SVM, RF, k-Nearest Neighbors (KNN), and AdaBoost. Experiments were conducted on the binary Breast\_cancer dataset and the multi-class Ecoli dataset. The relevant algorithms were implemented using the imbalanced-learn \cite{ref182} and smote-variants \cite{ref183} toolboxes. The parameter settings are all toolbox default values.

The results in Tables X and XI show that the ability of classifiers to model sample distribution and feature space significantly outweighs the optimization gained from resampling methods. In binary (Breast\_cancer) and multi-class (Ecoli) datasets, performance variations across resampling methods are minimal when the classifier is held constant. For example, the F1-score fluctuation with RF on breast\_cancer is only 0.5\%. However, classifier performance differences are more pronounced when the resampling method is fixed. For example,  the SVM classifier achieves 29.7\% higher accuracy than AdaBoost on the Ecoli dataset when using SMOTE. The effectiveness of resampling methods is also task-dependent. While SMOTE-IPF boosts AdaBoost’s F1-score to 98.25\% on breast\_cancer, its performance declines on Ecoli. Borderline-SMOTE enhances the G-mean of RF on Ecoli, but reduces the performance of DT on breast\_cancer. Methods such as SMOTE-ENN and ADASYN are less stable and may introduce noise or overfitting risks, requiring careful use based on the data distribution. Notably, AdaBoost with SMOTE-IPF performs best on breast\_cancer, while SVM retains better performance on Ecoli with the original distribution. In practice, classifier selection should be prioritized, followed by targeted resampling optimization, with careful consideration of data characteristics when selecting augmentation strategies.

In applications, it is best to choose a suitable classifier based on task needs rather than relying too heavily on sampling, to avoid overlooking the model's inherent capabilities. Sampling methods should match data characteristics. We should prioritize simple and interpretable strategies to minimize risks. During initial experiments, comparing the stability of different strategies and their synergy with the classifier is essential. Cross-validation can help ensure robustness by minimizing randomness. For certain tasks, preserving the original data distribution may be more effective than resampling, especially when the data is balanced or the classifier is sensitive to noise. At the same time, hybrid methods reveal the complementary relationship between the algorithm, data, and features. Therefore, the experiment highlights the importance of analyzing sample and feature distributions. The most effective method is one that aligns algorithm capabilities with dataset characteristics.

\begin{table*}
	\begin{center}
		\caption{Experimental Results on the Multi-class Ecoli Dataset.}
		\label{tab11}
		\begin{tabular}{ l|ccccc| ccccc }
			\hline	
			\textbf{Evaluation}& \textbf{Accuracy} &&&&  &\textbf{AUC}&&&&\\
			\hline								
			Method&	DT&	SVM&	RF&	KNN&	AdaBoost&
			DT&	SVM&	RF&	KNN&	AdaBoost\\
			\hline
			No sampling&	0.7624&	\textbf{0.8812}&	0.8218&	\textbf{0.8812}&	\textbf{0.6535}&
			0.6967&	\textbf{0.9141}&	0.8386&	0.8577&	0.6860\\
			SMOTE&	0.7921&	0.8812&	0.8713&	0.8416&	0.5842&
			0.7959&	0.8729&	0.8428&	0.8444&	0.7031\\
			Borderline\_SMOTE&	0.7921&	0.8614&	0.8515&	0.8416&	0.5446&
			0.7300&	0.8722&	0.8378&	0.8508&	0.5995\\
			ADASYN&	0.7426&	0.8713&	\textbf{0.8812}&	0.8119&	0.5743&
			0.7248&	0.7248&	0.7248&	0.7248&	0.7248\\
			A\_SUWO&	0.7921&	0.8317&	0.8713&	0.8614&	0.6535&
			0.7862&	0.8876&	0.8415&	0.8512&	0.5912\\
			MWMOTE&	0.7822&	0.8812&	0.8713&	0.8317&	0.2970&
			0.7849&	0.8767&	\textbf{0.8480}&	0.8466&	0.6857\\
			CCR&	0.7426&	0.8416&	0.8614&	0.8812&	0.6337&
			0.7645&	0.8945&	0.8435&	\textbf{0.8580}&	0.6607\\
			SMOTE\_TomekLinks&	0.7525&	0.8713&	0.8416&	0.8317&	0.6535&
			0.7877&	0.8737&	0.8471&	0.8431&	0.7031\\
			SMOTE-ENN&	0.8020&	0.8713&	0.8515&	0.8515&	0.6535&
			0.794&	0.8751&	0.8405&	0.8438&	0.7031\\
			SMOTE-IPF&	0.7921&	0.8812&	0.8713&	0.8416&	0.5842&
			0.7959&	0.8729&	0.8428&	0.8444&	0.7031\\
			MDO&	\textbf{0.8119}&	0.8812&	0.8218&	0.8515&	0.5941&
			\textbf{0.7963}&	0.8805&	0.8464&	0.8511&	\textbf{0.7313}\\
			\hline
			\hline
			\textbf{Evaluation}& \textbf{F1-score}&&&&  &\textbf{G-Mean}&&&&\\
			\hline
			Method&	DT&	SVM&	RF&	KNN&	AdaBoost&
			DT&	SVM&	RF&	KNN&	AdaBoost\\
			\hline
			no sampling&	0.7696&	0.872&	0.8167&	\textbf{0.8704}&	\textbf{0.5365}&
			0.8502&	0.9205&	0.884&	\textbf{0.9198}&	\textbf{0.7313}\\
			SMOTE&	0.7942&	\textbf{0.8732}&	0.8669&	0.8332&	0.4972&
			0.8722&	0.9223&	0.9191&	0.8985&	0.6982\\
			Borderline\_SMOTE&	0.7883&	0.8542&	0.8476&	0.8319&	0.4287&
			0.8673&	0.9100&	0.9067&	0.8968&	0.6603\\
			ADASYN&	0.7373&	0.8653&	\textbf{0.8753}&	0.8103&	0.4816&
			0.8271&	0.9221&	\textbf{0.9266}&	0.8853&	0.6810\\
			A\_SUWO&	0.7918&	0.8221&	0.8657&	0.8512&	0.5297&
			0.8680&	0.8899&	0.9191&	0.9086&	0.7213\\
			MWMOTE&	0.7829&	0.8705&	0.8659&	0.8238&	0.2106&
			0.8625&	\textbf{0.9256}&	0.9209&	0.8951&	0.5252\\
			CCR&	0.7442&	0.8336&	0.8561&	0.8691&	0.5179&
			0.8378&	0.9035&	0.9134&	0.9195&	0.7030\\
			SMOTE\_TomekLinks&	0.7524&	0.8618&	0.8372&	0.8242&	0.5365&
			0.8432&	0.9136&	0.9027&	0.8919&	0.7313\\
			SMOTE-ENN&	0.7986&	0.8618&	0.8483&	0.8463&	0.5365&
			0.8756&	0.9136&	0.9087&	0.9064&	0.7313\\
			SMOTE-IPF&	0.7942&	\textbf{0.8732}&	0.8669&	0.8332&	0.4972&
			0.8722&	0.9223&	0.9191&	0.8985&	0.6982\\
			MDO	&\textbf{0.8072}&	0.8725&	0.8165&	0.8432&	0.5304&
			\textbf{0.8805}&	0.9232&	0.8865&	0.9039&	0.6984\\
			\hline
		\end{tabular}
	\end{center}
\hspace{1cm}
\footnotesize \textit{Bolded values are the best among different sampling methods for the same classifier with the same evaluation metrics.}
\end{table*}

\section{Open Questions and Future Trends}
\textit{Comprehensive consideration of S\&I problem:} Effectively addressing S\&I requires understanding its root causes and identifying key factors influencing classification. While techniques such as data augmentation and balancing can improve accuracy, they often overlook data complexity. Research should analyze the internal structure of minority-class samples, identify instances with the greatest impact on the classifier, and optimize high-difficulty samples to enhance model performance \cite{ref184}. A deeper investigation into the interplay between class imbalance and other factors, guided by our analytical framework, can lead to more effective solutions.

\textit{Dataset Construction and Domain Adaptation:} How do synthetic and real data impact model performance differently? How can these effects be quantified? How can expanding the most effective sample size enhance the model's generalization? Data augmentation enhances training set size and minority-class diversity, but ensuring the authenticity and representativeness of generated samples remains a key challenge. Future research should focus on transfer learning, especially cross-domain knowledge transfer, and developing methods to quantify its effectiveness.

\textit{Algorithm Innovation and Adaptation:} Different learning algorithms vary in sensitivity to class imbalance \cite{ref185}. Adaptive methods to dynamically adjust weights and ensure generalization under extreme imbalance are crucial. In multi-class settings, exploring adaptive neighborhood size, multi-class combinatorial oversampling, and integrating class structure information into preprocessing and classifier design can improve handling \cite{ref62}. Key questions include: Which features most influence model performance, and how can feature selection and weighting be optimized for accuracy? In addition, as the demand for model transparency continues to grow, developing interpretable models will become increasingly important.

\textit{Applications of Few-Shot Learning:} Future research can reformulate S\&I as a Few-Shot or Zero-Shot Learning problem, focusing on improving model generalization and learning efficiency. Ji et al. \cite{ref186} proposed a semantic-guided class-imbalance learning model (SCILM), which mitigates the impact of class imbalance on Zero-Shot Image Classification through balanced training strategies and semantic-guided feature fusion. This approach also offers new insights for addressing the S\&I problem.

\textit{Multimodal Learning and Large Models:} How can minority-class samples from different modalities (e.g., image, text) be effectively integrated to improve overall performance? Future research should focus on leveraging the complementary information across modalities and developing effective fusion mechanisms to optimize learning outcomes. The advantages of large models can be harnessed to develop high-precision, task-specific methods. However, amid the rapid development of large-scale models, fundamental research and systematic methodological studies remain highly valuable and warrant further in-depth exploration.

\textit{Evaluation Metrics:} Current evaluation metrics mainly emphasize classification performance, often neglecting model interpretability assessment. Future work will explore metrics for model uncertainty, task-adaptive and dynamic evaluation, joint optimization of algorithms and metrics, and transferable evaluation methods across tasks and domains.

\section{Conclusion}
This paper comprehensively explores the S\&I problem. We advocate for a deeper investigation of S\&I datasets by leveraging feature analysis and imbalance metrics to identify and quantify factors that affect learning. The paper emphasizes the importance of selecting the most appropriate method based on the specific characteristics of the dataset, particularly one that strikes an optimal balance between cost and performance. The study clarifies the approach to addressing S\&I and offers valuable insights and directions for future research. While new technologies will likely lead to better solutions, the ideas presented in this paper remain a valuable reference.


\vfill

\end{document}